%% file: main.tex
\documentclass[10pt]{article} 

\usepackage[preprint]{tmlr}

\usepackage[utf8]{inputenc} 
\usepackage{url}            
\usepackage{booktabs}       
\usepackage{array}          
\usepackage{amsmath}        
\usepackage{amsthm}         
\usepackage{amsfonts}       
\usepackage{amssymb}        
\usepackage{nicefrac}       
\usepackage{microtype}      
\usepackage{xcolor}         
\usepackage{graphicx}       
\usepackage{wrapfig}        
\usepackage[bottom]{footmisc} 
\usepackage{fvextra}        
\usepackage{tcolorbox}
\usepackage{tikz}           
\usetikzlibrary{shapes.geometric}
\usepackage[
    colorlinks=true,
    linkcolor=blue!50!black,
    citecolor=blue!50!black,
    urlcolor=blue!50!black,
    pdfborder={0 0 0},
    pdfhighlight=/N
]{hyperref}

\usepackage{enumitem}
\usepackage{needspace}
\usepackage{placeins}
\usepackage{etoolbox}
\preto\subsection{\Needspace*{4\baselineskip}}
\preto\subsubsection{\Needspace*{4\baselineskip}}
\preto\paragraph{\Needspace*{4\baselineskip}}

\newcommand{\TODO}[1]{\textcolor{black}{\textnormal{#1}}}
\newcommand{\TODOcolor}{\color{black}}  

\title{Prior Knowledge or Search? A Study of LLM Agents in\\Hardware-Aware Code Optimization}

\newcommand{\authorcell}[3]{%
  \begin{tabular}[t]{@{}c@{}}
    {\large\bf #1}\\[2pt]
    {\normalfont #2}\\[2pt]
    {\normalfont\texttt{\small #3}}
  \end{tabular}}

\author{\normalfont\hbox to \hsize{\hss
\begin{tabular}{@{}c@{\hspace{0.5cm}}c@{\hspace{0.5cm}}c@{}}
\authorcell{Dmitry Redko\thanks{Authors contributed equally to this work.}}{Applied AI Institute}{dmitryredko444@gmail.com} &
\authorcell{Albert Fazlyev\footnotemark[1]}{AI Talent Hub, ITMO University}{albert.fz@yandex.ru} &
\authorcell{Konstantin Sozykin\footnotemark[1]}{Applied AI Institute}{k.sozykin@applied-ai.ru} \\ \noalign{\vskip 0.22in}
\authorcell{Maria Ivanova}{YSDA, Applied AI Institute}{ivanova.m.pe@gmail.com} &
\authorcell{Evgeny Burnaev}{Applied AI Institute}{e.burnaev@applied-ai.ru} &
\authorcell{Egor Shvetsov}{Applied AI Institute}{e.shvetsov@applied-ai.ru}
\end{tabular}\hss}}

\def\month{MM}
\def\year{YYYY}
\def\openreview{\url{https://openreview.net/forum?id=XXXX}}

\begin{document}

\maketitle

\input{sections/00_abstract}
\input{sections/01_introduction}
\input{sections/03_preliminary}
\input{sections/04_bbo}
\input{sections/05_kernel_gen}
\input{sections/06_discussion_conclusion}

\bibliography{references}
\bibliographystyle{tmlr}

\clearpage
\appendix
\input{sections/99_appendix}

\end{document}

%% file: sections/00_abstract.tex
\begin{abstract}
LLM discovery and optimization systems are increasingly applied across domains, implementing a common propose-evaluate-revise loop. Such optimization or discovery progresses via context conditioning on feedback received from an environment. In classical optimization methods, the optimizer's search state is updated, while in LLM-based systems the context is updated but the weights stay fixed. In this work we show that the rich prior encoded in an LLM's weights can be useful in some scenarios, but misleading and detrimental when prior knowledge does not transfer to the task. To demonstrate this, we develop three controlled scenarios, one is optimization over unknown functions, and two low-level code optimization tasks. We find that (1) in function optimization, LLMs act as greedy optimizers and underperform classic methods. (2) In low-level code optimization, performance sharply degrades when generating kernels for uncommon sizes and providing explicit input-size information has no measurable effect. (3) Iterative generation degrades when a less common language is used. Compiled-and-correct rates generally increase for CUDA with iterations but decrease for TVM IR. (4) In several scenarios, classical optimization methods outperform LLM-based systems, even on code optimization tasks. Our results empirically highlight a core limitation of current LLM-based optimization and discovery systems as their performance depends more on pretrained knowledge than on the accumulated context and feedback.

\end{abstract}

%% file: sections/01_introduction.tex
\section{Introduction}
\label{sec:introduction}

Large language models are increasingly deployed as autonomous discovery systems that propose candidate solutions, observe evaluative feedback, and iterate toward novel results. Recent successes span extremal combinatorics~\citep{funsearch2024}, matrix multiplication~\citep{alphaevolve2025}, scientific equation discovery~\citep{shojaee2025llmsr}, chemistry~\citep{yang2025moosechem}, and GPU kernel optimization~\citep{ouyang2025kernelbench,lange2025cudaengineer} all sharing the same high level structure, a \textit{propose} $\rightarrow$ \textit{evaluate} $\rightarrow$ \textit{revise} loop. This pattern is structurally identical to Black-Box Optimization (BBO)~\citep{shahriari2015taking}, with one key difference: a surrogate model updates a posterior over the objective landscape, whereas the LLM encodes its prior in frozen weights and adapts solely through context conditioning. LLMs are effective when this frozen-weight prior, or captured knowledge, supports the target solution space. However, we hypothesize that neither the ability to reason out a solution nor the ability to explore for one will compensate for the lack of a relevant prior in an unfamiliar domain. Although this hypothesis seems simple, it is hard to assess empirically, because prior alignment is not directly observable, and in full agentic pipelines its effect is hard to separate from scaffolding, feedback quality, and search budget. To investigate it, we develop three controlled settings: one involving unknown functions, where task-specific knowledge is limited, and two involving more practical, low-level code (kernel) optimization tasks. In kernel optimization, we control prior density along two axes: (1) representation language, comparing CUDA with TVM intermediate representation (IR), which we reasonably assume is less represented in the models' training data\footnote{\small This assumption is based on TVM being less frequent in The Stack~\citep{kocetkov2022thestack}, and is further supported by our analysis of the models' output entropy across programming languages in Appendix~\ref{sec:appendix-entropy}.}, and (2) solution-space familiarity, contrasting standard kernel sizes likely seen during pretraining with small, atypical sizes that are rarely encountered.

\needspace{5\baselineskip}
\textbf{We summarize our findings and contributions as follows.}
\begin{enumerate}[leftmargin=*]

\item \textbf{Empirical evidence that LLMs perform greedy optimization and
require orchestrated exploration to overcome it~(\S\ref{sec:setup-bbo}).}
LLMs act as greedy optimizers when optimizing unknown functions, losing to classical algorithms
on tasks that require genuine exploration. Combining LLMs with a method that explicitly encourages exploration improves performance.

\item \textbf{Weak programming-language priors degrade iterative feedback-loop exploration~(\S\ref{sec:setup-kernel}).}
The quality of the LLM's prior over the target language determines whether
iterative optimization converges or deteriorates. More precisely, the number of correctly compiled kernels generally increases across iterations for CUDA but decreases at each step for TVM IR, where
the prior is too weak to support the feedback loop.

\item \textbf{Performance degrades sharply when generating kernels for uncommon sizes~(\S\ref{sec:setup-kernel} and \S\ref{sec:zero-shot-prior}).} Moreover, providing explicit input-size information has no measurable effect as models generate CUDA kernels with generic parameters regardless of the requested size.

\item \textbf{Classical methods outperform the LLM where the prior is thin~(\S\ref{sec:setup-bbo} and \S\ref{sec:setup-kernel}).} At a fixed evaluation budget, CMA-ES and its LLM
hybrid Centaur~\citep{ferreira2026llmsbeatclassicalhyperparameter} outperform
the plain LLM on the task families that require exploration, and TVM MetaSchedule\footnote{TVM MetaSchedule uses XGBoost to score mutated kernel variants and selects a subset to evaluate on hardware; this process is repeated over several iterations.} wins on roughly $70\%$ of kernels when the input
sizes move away from the benchmark's canonical regime.

\end{enumerate}

\textbf{Practical implications.} Modern agentic pipelines combine context manipulation, iterative feedback, population-based evolution, and randomized search in ways that make it difficult to attribute performance to any single factor. By working in controlled, minimal settings we isolate these factors and expose failure modes fundamental to any LLM-based discovery system, with direct implications for the design of LLM agents in general and code optimization agents in particular.

\textbf{Paper roadmap.} \S\ref{sec:framework} frames LLM discovery as a form of
black-box optimization with a frozen, context-conditioned prior. \S\ref{sec:setup-bbo}
tests this view with known functions optimization, where no strong code prior is available, and demonstrates 
that LLMs behave as greedy local refiners. \S\ref{sec:results} then moves
to kernel optimization, where we vary prior alignment through input shape,
programming languages, and feedback architecture. Finally,
\S\ref{sec:discussion} interprets these results and discusses when LLM
agents should be paired with explicit search or retrieval rather than used as
standalone optimizers. Related work and additional experimental details and
results are deferred to the appendix.

%% file: sections/03_preliminary.tex
\section{A Conceptual View of LLM Agents vs. classic BBO}
\label{sec:framework}

We compare LLM agents and classic black-box optimization (BBO) methods in the same loop.
Both propose a candidate, evaluate it, and use the result to choose the next
one. For a candidate $x$ in the search space $\mathcal{X}$, at step $k$ the method holds its own parameters $\theta_k$ and a context $c_k$
and produces the next candidate,
\begin{align}
x_{k+1} = q(\theta_k, c_k).
\end{align}
Here $\theta_k$ are the parameters of the method, the search state for
BBO and the model weights for an LLM. Each candidate is evaluated and the
environment returns a signal $y_{k+1}$, which can be a scalar objective value,
a compiler error, or a profiling report. The context $c_k$ collects the
previous candidates and their signals. In both cases $q$ is a function that
returns the next candidate.

\textbf{Black-Box Optimization.} Classic BBO methods keep no explicit context.  Each
new signal is folded into the parameters by an update rule, so the history
reaches the next proposal only through $\theta_k$,
\begin{align}
x_{k+1} = q(\theta_k), \qquad
\theta_{k+1} = U(\theta_k, x_{k+1}, y_{k+1}).
\end{align}
The parameters $\theta_k$ form a parametric model of the task that is
fitted from the observations collected over the iterations of the current run,
using an update rule $U$ set by the algorithm. The initial $\theta_0$ carries
no task information. The classical example is Bayesian optimization,
where $\theta_k$ parameterizes a surrogate with mean $\mu_k$ and uncertainty
$\sigma_k$ and the next point maximizes an acquisition function such as
$\alpha_{\mathrm{UCB}}(x) = \mu_k(x) + \kappa\sigma_k(x)$
~\citep{snoek2012practical}. Within this same scheme, CMA-ES instead sets $\theta_k$ to the mean, step size, and covariance of a Gaussian search distribution~\citep{cma_es_tutorial}.

\textbf{LLM Agent.}  The parameters are
the same object, but they are frozen, and every new signal is captured in the
context,
\begin{align}
x_{k+1} = q(\theta, c_k), \qquad
c_{k+1} = \pi(c_k, x_{k+1}, y_{k+1}),
\end{align}
where $\pi$ is the agent scaffold. Because $\theta$ does not change
over the iterations of a run, no task-specific quantity is fitted. These frozen weights
are what we call \emph{prior knowledge}. They aggregate everything the model
acquired during pretraining and post-training, and they alone decide how the
context is turned into a candidate. Feedback enters only through $c_k$, so it
can change what the model sees but not how the model reads it. The prior is
strong when the task representation and typical solutions were common in
training, and weak otherwise. Prompts, history, feedback, tool outputs, and
search state belong to $c_k$ and are not prior knowledge.

\textbf{Practical implication.} Classical BBO methods adapt to any task but start from
scratch each time. An LLM agent can make good guesses before much feedback is
available, but it cannot change how it interprets that feedback. The
next sections test where the LLM's prior helps and where it fails.

%% file: sections/04_bbo.tex
\section{LLMs as Black-Box Optimizers}
\label{sec:setup-bbo}
The model is never told which function or system it is optimizing: it
sees only the search bounds and the history of its own proposals and their
losses. It does, however, recognize that it is performing optimization.

\subsection{Methods and Evaluation for BBO Problems}

\textbf{Problems.} We evaluate on four task sets, three two-dimensional and one five-dimensional. 
The first set contains 100 2d synthetic problems with unknown closed-form 
\textbf{Functions} that have multiple local minima. The second set contains 
100 2d problems based on equations from \textbf{Physical} dynamical systems.  Both task sets were manually constructed. Functions are random
compositions of elementary functions, whereas Physical tasks are based on
physical laws with time-dependent perturbations. Their construction is
described in Appendix~\ref{app:functions-physical-construction}.
The third and fourth sets come from the \textbf{BBOB} benchmark~\citep{bbob2009}
and contain 48 2d and 48 5d problems, respectively.

\textbf{Optimizers.}
For each problem, we evaluate \textbf{four optimizers} under the same 50-trial
budget.

\emph{(1) Pure LLM} is a sequential black-box proposer. At each
step, it receives only the search history containing previous proposals,
corresponding values, and the task prompt, and returns the next candidate
point.

\emph{(2) CMA-ES}~\citep{cma_es_tutorial}, implemented via Optuna~\citep{akiba2019optuna} and chosen for its interpretable Gaussian state (mean, step-size, covariance), adapts from objective feedback.

\emph{(3) Centaur}~\cite{ferreira2026llmsbeatclassicalhyperparameter} enables true co‑adaptation by letting the LLM override CMA‑ES proposals while CMA‑ES learns from them, unlike other hybrids~\citep{liu2024large}.

\emph{(4) LLM-MCTS} wraps the LLM proposer in a Monte Carlo Tree Search~\citep{coulom2006efficient} scaffold. Each expansion selects a node via UCB1~\citep{kocsis2006bandit}, prompts the LLM with the root-to-node path as context, samples $k$ candidate points in parallel, attaches all $k$ as children of the selected node, and back-propagates the minimum child loss through the ancestor chain.

For pure LLM and hybrid methods, we evaluate \texttt{gpt-oss-120b},
\texttt{DeepSeek-V3.2} (685B), and \texttt{DeepSeek-V4-Pro} (1.6T) on the complete set of problems using identical prompts, optimizer
configurations, budgets, and metrics. Prompt templates for hybrid and pure LLMs experiments are provided in
Appendix~\ref{sec:appendix-bbo-prompts}.

\textbf{Evaluation metrics.}
We summarize optimizers behavior over fixed budgets through three
complementary quantities. \emph{(1) Best step} is the trial at which the best
loss is first attained and measures convergence speed. \emph{(2) Coverage} is
the fraction of the search domain probed by the query sequence and measures
exploration breadth. \emph{(3) Normalized trajectory length} is the ratio of
end-to-end displacement to total path length and measures directional
consistency, with $L=1$ for a perfectly linear path and $L \approx 0.14$ for an
isotropic random walk at $K=50$. Full definitions are provided in
Appendix~\ref{app:metrics}.

\begin{figure*}[!t]
\centering
\includegraphics[width=0.85\linewidth]{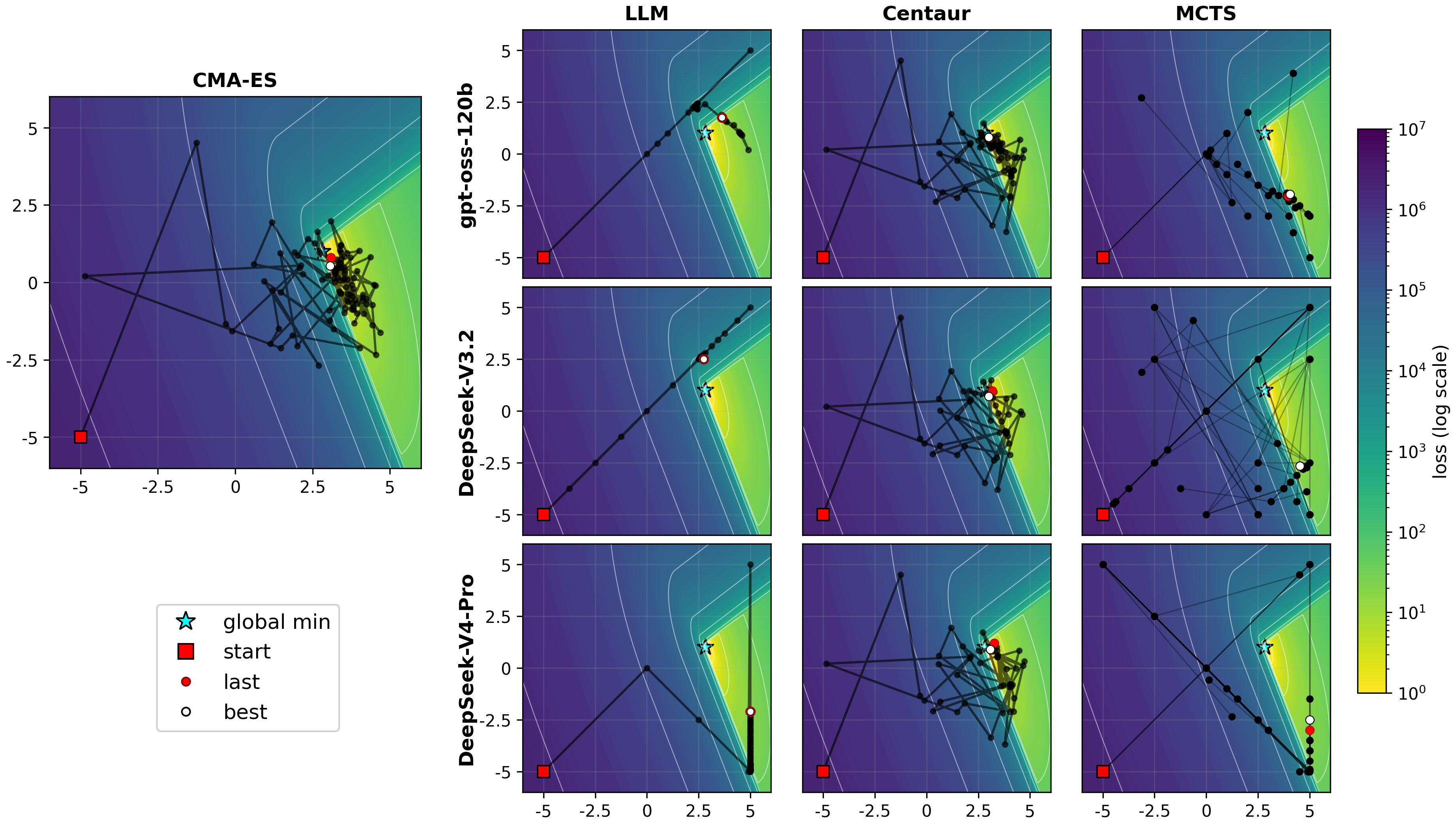}
\caption{An example of optimization traces.
An LLM often follows a greedy, line-like trajectory that either converges quickly or commits to a suboptimal direction. Additional illustrations of two dimensional BBO trajectories are presented in Appendix~\ref{sec:appendix-bbob-traces}.}
\label{fig:optuna-llm-trajectories}
\end{figure*}

\begin{table*}[!t]
\centering
\setlength{\tabcolsep}{3.2pt}
\renewcommand{\arraystretch}{1.12}
\caption{Results for BBO optimization. Entries in the LLM vs optimizer block are pairwise win counts in the form
plain-LLM wins vs method wins, with the larger count shown in bold. Each following block presents different metric for each method. The methods are LLM, CMA-ES (CME), Centaur (Cent.),
and MCTS. The metrics are average best step, final coverage~$\mathrm{Cov}_{50}$ (\%),
and normalized trajectory length~$L$, all definitions are provided in
Appendix~\ref{app:metrics}. }

\resizebox{\textwidth}{!}{%
\begin{tabular}{@{}ll|ccc|cccc|cccc|cccc@{}}
\toprule
\textbf{Model}
& \textbf{Task}
& \multicolumn{3}{c|}{\textbf{LLM vs optimizer}}
& \multicolumn{4}{c|}{\textbf{Avg.\ best step}}
& \multicolumn{4}{c|}{\textbf{Coverage (\%)}}
& \multicolumn{4}{c}{\textbf{$L$}} \\
\cmidrule(lr){3-5}\cmidrule(lr){6-9}\cmidrule(lr){10-13}\cmidrule(lr){14-17}
&
& \textbf{CME}
& \textbf{Cent.}
& \textbf{MCTS}
& \textbf{LLM}
& \textbf{CME}
& \textbf{Cent.}
& \textbf{MCTS}
& \textbf{LLM}
& \textbf{CME}
& \textbf{Cent.}
& \textbf{MCTS}
& \textbf{LLM}
& \textbf{CME}
& \textbf{Cent.}
& \textbf{MCTS} \\
\midrule
\texttt{gpt-oss-120b} & Functions & $\mathbf{78}\,\mathrm{vs}\,22$ & $\mathbf{75}\,\mathrm{vs}\,25$ & $\mathbf{83}\,\mathrm{vs}\,17$ & $42.8$ & $31.8$ & $37.7$ & $\mathbf{27.6}$ & $10.9$ & $44.8$ & $45.3$ & $\mathbf{93.4}$ & $0.72$ & $\mathbf{0.16}$ & $\mathbf{0.16}$ & $\mathbf{0.16}$ \\
& Physical & $37\,\mathrm{vs}\,\mathbf{63}$ & $29\,\mathrm{vs}\,\mathbf{71}$ & $49\,\mathrm{vs}\,\mathbf{51}$ & $46.7$ & $35.3$ & $44.1$ & $\mathbf{33.4}$ & $22.4$ & $71.2$ & $71.4$ & $\mathbf{93.9}$ & $0.67$ & $\mathbf{0.17}$ & $0.19$ & $\mathbf{0.17}$ \\
& BBOB & $23\,\mathrm{vs}\,\mathbf{25}$ & $18\,\mathrm{vs}\,\mathbf{30}$ & $\mathbf{25}\,\mathrm{vs}\,23$ & $40.0$ & $36.2$ & $39.1$ & $\mathbf{36.1}$ & $11.7$ & $37.1$ & $33.1$ & $\mathbf{82.1}$ & $0.68$ & $\mathbf{0.18}$ & $\mathbf{0.18}$ & $0.27$ \\
& BBOB-5D & $11\,\mathrm{vs}\,\mathbf{37}$ & $11\,\mathrm{vs}\,\mathbf{37}$ & $19\,\mathrm{vs}\,\mathbf{29}$ & $46.7$ & $\mathbf{34.7}$ & $43.6$ & $42.3$ & $11.8$ & $41.1$ & $48.5$ & $\mathbf{69.8}$ & $0.60$ & $\mathbf{0.11}$ & $0.14$ & $0.32$ \\
\addlinespace[2pt]
\texttt{DeepSeek-V3.2} & Functions & $\mathbf{80}\,\mathrm{vs}\,20$ & $\mathbf{81}\,\mathrm{vs}\,19$ & $\mathbf{88}\,\mathrm{vs}\,12$ & $40.7$ & $31.8$ & $37.5$ & $\mathbf{30.4}$ & $24.1$ & $44.8$ & $41.8$ & $\mathbf{97.1}$ & $0.61$ & $0.16$ & $0.16$ & $\mathbf{0.15}$ \\
& Physical & $42\,\mathrm{vs}\,\mathbf{58}$ & $36\,\mathrm{vs}\,\mathbf{64}$ & $\mathbf{65}\,\mathrm{vs}\,35$ & $43.3$ & $35.3$ & $42.6$ & $\mathbf{30.3}$ & $28.0$ & $71.2$ & $68.8$ & $\mathbf{99.3}$ & $0.63$ & $0.17$ & $0.21$ & $\mathbf{0.13}$ \\
& BBOB & $\mathbf{28}\,\mathrm{vs}\,20$ & $\mathbf{25}\,\mathrm{vs}\,23$ & $\mathbf{33}\,\mathrm{vs}\,15$ & $35.8$ & $36.2$ & $38.0$ & $\mathbf{31.6}$ & $33.4$ & $37.1$ & $35.7$ & $\mathbf{97.2}$ & $0.57$ & $0.18$ & $\mathbf{0.17}$ & $0.19$ \\
& BBOB-5D & $12\,\mathrm{vs}\,\mathbf{36}$ & $9\,\mathrm{vs}\,\mathbf{39}$ & $23\,\mathrm{vs}\,\mathbf{25}$ & $36.2$ & $34.7$ & $41.3$ & $\mathbf{24.7}$ & $17.3$ & $41.1$ & $51.8$ & $\mathbf{97.9}$ & $0.42$ & $\mathbf{0.11}$ & $0.12$ & $0.14$ \\
\addlinespace[2pt]
\TODO{\texttt{DeepSeek-V4-Pro}} & \TODO{Functions} & \TODO{$\mathbf{89}\,\mathrm{vs}\,11$} & \TODO{$\mathbf{88}\,\mathrm{vs}\,12$} & \TODO{$\mathbf{90}\,\mathrm{vs}\,10$} & \TODO{$36.7$} & \TODO{$\mathbf{31.8}$} & \TODO{$42.8$} & \TODO{$32.2$} & \TODO{$70.9$} & \TODO{$44.8$} & \TODO{$50.0$} & \TODO{$\mathbf{98.8}$} & \TODO{$0.53$} & \TODO{$0.16$} & \TODO{$\mathbf{0.15}$} & \TODO{$\mathbf{0.15}$} \\
& \TODO{Physical} & \TODO{$40\,\mathrm{vs}\,\mathbf{60}$} & \TODO{$35\,\mathrm{vs}\,\mathbf{65}$} & \TODO{$47\,\mathrm{vs}\,\mathbf{53}$} & \TODO{$47.9$} & \TODO{$35.3$} & \TODO{$45.8$} & \TODO{$\mathbf{34.3}$} & \TODO{$14.4$} & \TODO{$71.2$} & \TODO{$69.9$} & \TODO{$\mathbf{95.8}$} & \TODO{$0.75$} & \TODO{$\mathbf{0.17}$} & \TODO{$0.20$} & \TODO{$\mathbf{0.17}$} \\
& \TODO{BBOB} & \TODO{$\mathbf{26}\,\mathrm{vs}\,22$} & \TODO{$\mathbf{25}\,\mathrm{vs}\,23$} & \TODO{$\mathbf{28}\,\mathrm{vs}\,20$} & \TODO{$41.5$} & \TODO{$36.2$} & \TODO{$46.0$} & \TODO{$\mathbf{35.3}$} & \TODO{$33.6$} & \TODO{$37.1$} & \TODO{$35.1$} & \TODO{$\mathbf{70.8}$} & \TODO{$0.72$} & \TODO{$\mathbf{0.18}$} & \TODO{$\mathbf{0.18}$} & \TODO{$0.23$} \\
& \TODO{BBOB-5D} & \TODO{$16\,\mathrm{vs}\,\mathbf{32}$} & \TODO{$11\,\mathrm{vs}\,\mathbf{37}$} & \TODO{$\mathbf{32}\,\mathrm{vs}\,16$} & \TODO{$47.2$} & \TODO{$34.7$} & \TODO{$43.3$} & \TODO{$\mathbf{32.7}$} & \TODO{$29.8$} & \TODO{$41.1$} & \TODO{$49.1$} & \TODO{$\mathbf{68.7}$} & \TODO{$0.56$} & \TODO{$\mathbf{0.11}$} & \TODO{$0.13$} & \TODO{$0.22$} \\
\midrule
\TODO{\textbf{Average}} & \TODO{Functions} & \TODO{$\mathbf{82}\,\mathrm{vs}\,18$} & \TODO{$\mathbf{81}\,\mathrm{vs}\,19$} & \TODO{$\mathbf{87}\,\mathrm{vs}\,13$} & \TODO{$40.1$} & \TODO{$31.8$} & \TODO{$39.3$} & \TODO{$\mathbf{30.1}$} & \TODO{$35.3$} & \TODO{$44.8$} & \TODO{$45.7$} & \TODO{$\mathbf{96.4}$} & \TODO{$0.62$} & \TODO{$0.16$} & \TODO{$0.16$} & \TODO{$\mathbf{0.15}$} \\
 & \TODO{Physical} & \TODO{$40\,\mathrm{vs}\,\mathbf{60}$} & \TODO{$33\,\mathrm{vs}\,\mathbf{67}$} & \TODO{$\mathbf{54}\,\mathrm{vs}\,46$} & \TODO{$46.0$} & \TODO{$35.3$} & \TODO{$44.2$} & \TODO{$\mathbf{32.7}$} & \TODO{$21.6$} & \TODO{$71.2$} & \TODO{$70.0$} & \TODO{$\mathbf{96.3}$} & \TODO{$0.68$} & \TODO{$0.17$} & \TODO{$0.20$} & \TODO{$\mathbf{0.16}$} \\
 & \TODO{BBOB} & \TODO{$\mathbf{26}\,\mathrm{vs}\,22$} & \TODO{$23\,\mathrm{vs}\,\mathbf{25}$} & \TODO{$\mathbf{29}\,\mathrm{vs}\,19$} & \TODO{$39.1$} & \TODO{$36.2$} & \TODO{$41.0$} & \TODO{$\mathbf{34.3}$} & \TODO{$26.2$} & \TODO{$37.1$} & \TODO{$34.6$} & \TODO{$\mathbf{83.4}$} & \TODO{$0.66$} & \TODO{$\mathbf{0.18}$} & \TODO{$\mathbf{0.18}$} & \TODO{$0.23$} \\
 & \TODO{BBOB-5D} & \TODO{$13\,\mathrm{vs}\,\mathbf{35}$} & \TODO{$10\,\mathrm{vs}\,\mathbf{38}$} & \TODO{$\mathbf{25}\,\mathrm{vs}\,23$} & \TODO{$43.4$} & \TODO{$34.7$} & \TODO{$42.7$} & \TODO{$\mathbf{33.2}$} & \TODO{$19.6$} & \TODO{$41.1$} & \TODO{$49.8$} & \TODO{$\mathbf{78.8}$} & \TODO{$0.53$} & \TODO{$\mathbf{0.11}$} & \TODO{$0.13$} & \TODO{$0.23$} \\
\bottomrule
\end{tabular}
}
\label{tab:optuna-llm-summary}
\end{table*}

\begin{figure}[!t]
\centering
\includegraphics[width=0.6\linewidth]{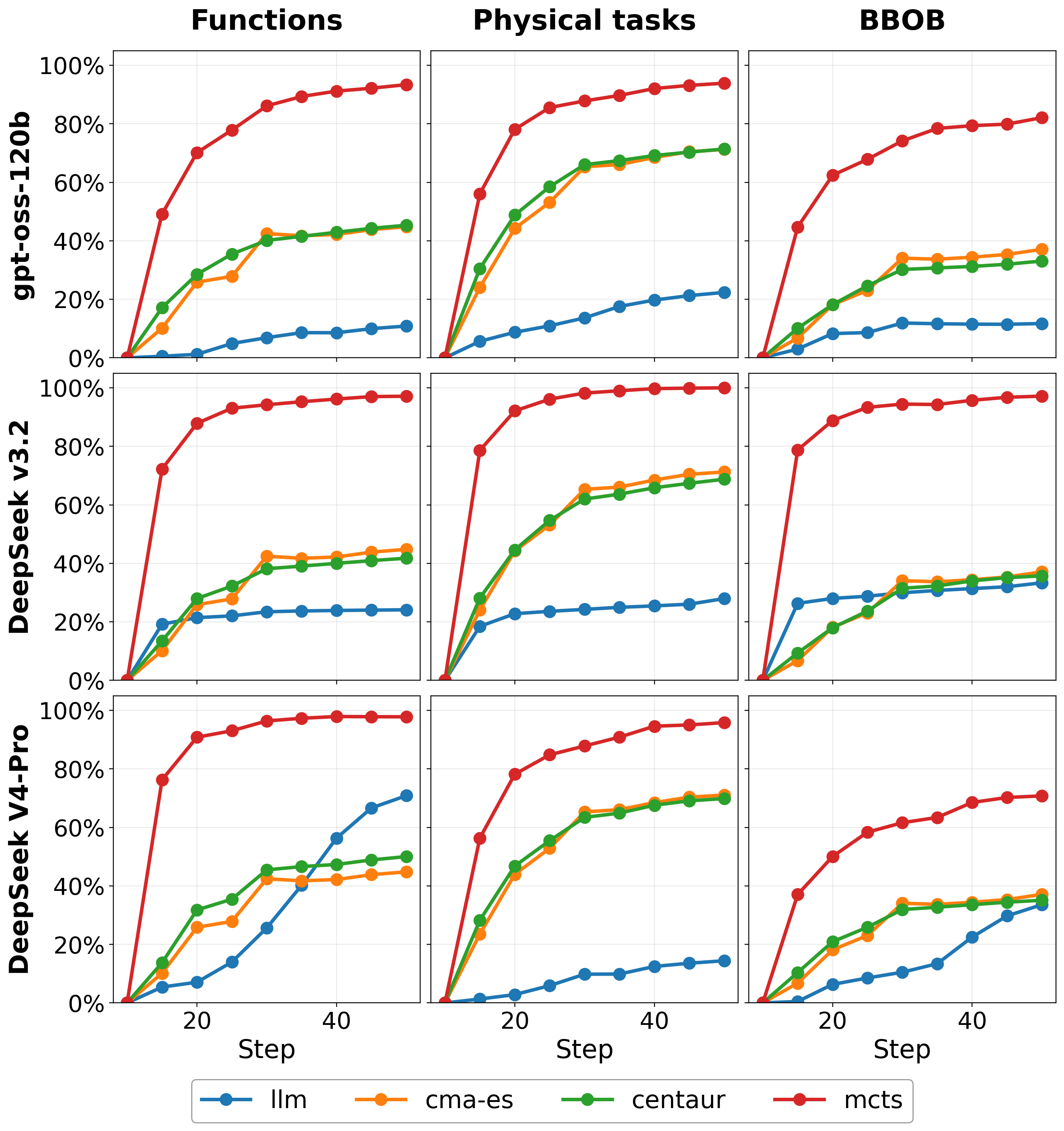}
\caption{Per-step search-space coverage $\mathrm{Cov}_k$, measured every five steps over 50 trials. The LLM has the lowest coverage. CMA-ES and Centaur occupy a similar mid-range, while LLM-MCTS
achieves the highest coverage in every task.}
\label{fig:bbo-coverage-dynamics}
\end{figure}

\subsection{Results for LLMs as BBO Optimizer}
\label{sec:llm-vs-optuna}

Table~\ref{tab:optuna-llm-summary} reports pairwise win counts, average 
best step, coverage $\mathrm{Cov}_{50}$, and normalized trajectory length 
$L$. Figure~\ref{fig:bbo-coverage-dynamics} shows coverage over steps. 
Together, they reveal three distinct exploration regimes.

\textbf{(1) LLMs act greedily, which suffices when the objective has many
local minima but breaks when the optimum is isolated.} LLMs win on functions, where many local minima make
greedy descent toward any plausible basin sufficient. On physical
tasks, however, where optima are isolated and exploration is necessary,
all three models underperform CMA-ES and Centaur. Low \textit{Coverage} and high \textbf{$L$} indicate
long, locally-confined trajectories rather than broad search. On \textit{BBOB-5D} all three LLMs lose to CMA-ES and Centaur.

\textbf{(2) MCTS over-explores.} It covers the domain
rapidly but exhausts its budget on exploration, rarely exploiting
promising regions though we note this is partly a consequence of our
hyperparameter choices.

\textbf{(3) CMA-ES and Centaur have the best balance.} They demonstrate  moderate
\textit{Coverage} and trajectory length achieving the
strongest average performance across tasks.

Appendix~\ref{app:bbob-budget-sensitivity}  presents results on \textit{BBOB-5D} 
at 75 and 100 trials - overall observations remain the same.

\begin{figure}[t]
\centering
\includegraphics[width=0.85\linewidth]{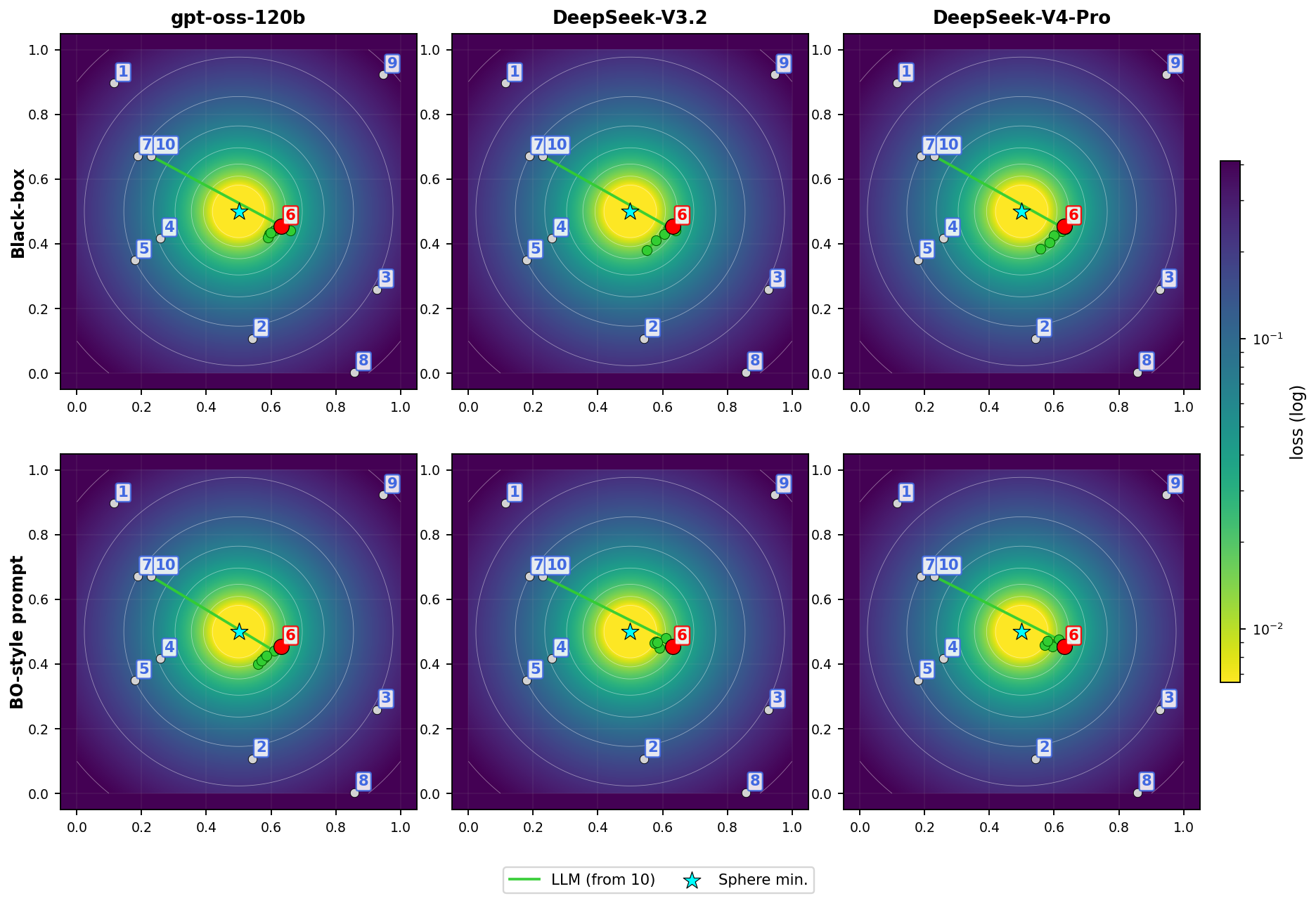}
\caption{\textit{Is the next step greedy?}
Steps~1-10 are uniform-random samples (white labeled circles), the best of
these is highlighted in red. The green segment marks future LLM-proposed steps~11-15
and the cyan star is the true minimum. LLMs next proposal is always close to
the best in history point regardless prompt.}
\label{fig:greedy-gradient-subplot}
\end{figure}

When prompted to act as Bayesian optimizers, LLMs still behaved greedily (Figure~\ref{fig:greedy-gradient-subplot}). Inspection of the models' chain-of-thought confirms that they attempt to mimic a gradient strategy to justify each successive proposal (Appendix~\ref{sec:appendix-greedy-gradient}, Table~\ref{tab:greedy-gradient-quotes}). We further investigate whether earlier steps influence model decisions or whether the model anchors solely on the current best result. To this end, we sampled 10 points on a two-dimensional sphere and then asked each model to propose 5 additional steps. In every case, the next proposal was closest to the current best solution rather than to the most recent step. Results are shown in Figure~\ref{fig:greedy-gradient-subplot}. Experimental protocols and prompt templates are provided in Appendix~\ref{sec:appendix-greedy-gradient}. 
\begin{tcolorbox}[
    colback=white,
    colframe=black,
    boxrule=0.5pt,
    top=1mm, bottom=1mm
]
With no prior knowledge of the
objective, the model falls back on the prior knowledge it does have, knowledge
of optimization as a subject, and the strategy it reaches for is local
refinement around the best observed point. We speculate this may also be a
consequence of the training objective, in which the model is rewarded for
predicting the single most likely next token. 
\end{tcolorbox}

%% file: sections/05_kernel_gen.tex
\section{Kernel Optimization}
\label{sec:results}

This section presents three findings on how the LLM's prior knowledge shapes kernel
generation:
\begin{enumerate}[label=(\roman*),leftmargin=*]
    \item Iterative feedback-loop optimization degrades rather than improves when prior knowledge is thin (\S~\ref{sec:setup-kernel}).
    \item In low-prior settings, classical methods outperform the LLM. TVM
    MetaSchedule beats the LLM agent on roughly $70\%$ of kernels when tasked with generating kernels for less common input shapes (\S~\ref{sec:setup-kernel}).
    \item Generated kernels reuse a generic set of kernel parameters, even when the prompt explicitly specifies the input shape. However, this generic set of parameters is not optimal for all input shapes (\S~\ref{sec:zero-shot-prior}).
\end{enumerate}

\subsection{Iterative Hardware-Aware Kernel Optimization with Feedback}
\label{sec:setup-kernel}
\begin{wrapfigure}{r}{0.45\linewidth}
\centering
\includegraphics[width=\linewidth]{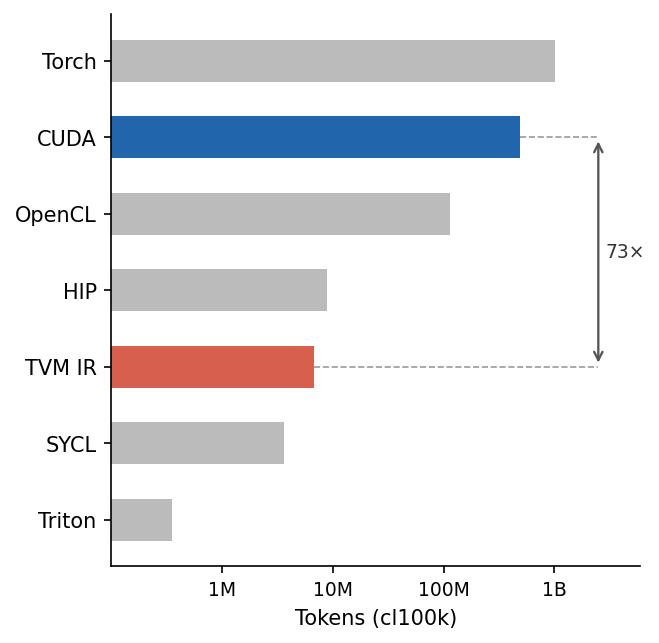}
\caption{Token counts for GPU-related languages in
The Stack. The x-axis is logarithmic.}
\label{fig:thestack-dist}
\end{wrapfigure}

We evaluate LLM-optimized kernels in terms of correctness and latency across two code representations (CUDA and TVM~IR), two optimization pipelines, and two input-size regimes. We first reasonably assume that TVM~IR is less represented than CUDA in the models' training data. Frequency in The Stack~\citep{kocetkov2022thestack} is a common proxy for how well a language is represented in pretraining data~\citep{cassano2024multiplt}, and Figure~\ref{fig:thestack-dist} shows that TVM is indeed less frequent than CUDA there. However, because public-code counts are only a proxy for what a model has seen, particularly for models with undisclosed training data, we also measure the token-level entropy of the models' own generations and find that it is higher for TVM~IR than for CUDA across all operations and models we tested (Appendix~\ref{sec:appendix-entropy}).

\textit{Procedure:} For CUDA, the model translates a PyTorch reference to a CUDA kernel.
For TVM~IR, the reference is an \emph{unscheduled} TVM-TIR program derived from
the same PyTorch model via automated conversion (ONNX$\to$Relax$\to$TIR, with
schedule annotations stripped), so the computation is already correctly
implemented. This allows us to have two computationally equivalent tasks but represented differently. The pipeline is described in
Appendix~\ref{sec:appendix-tvm-tir-reference}.  The TVM agent uses injected syntax guidance needed to reliably generate
valid code~\footnote{Our initial experiments demonstrated significantly lower performance without syntax guidance for TVM.}. Full prompts are in Appendix~\ref{sec:appendix-tvm-prompts}.

\textbf{Size regimes.} We evaluate two input-size regimes using problems from
KernelBench~\citep{ouyang2025kernelbench}. The \emph{Base} (benchmark-native) regime uses the original shapes and the
\emph{Small} regime uses substantially reduced sizes (e.g.\ matrix
multiplication shrinks from $M{=}2048,K{=}8192,N{=}4096$ to
$M{=}128,K{=}512,N{=}256$).
More examples are provided in Appendix~\ref{sec:appendix-size-table}.
All experiments cover two kernel difficulty levels of KernelBench with 100 kernels
per level for each size regime. 

\textbf{Baselines.}
\texttt{torch.compile} is the main performance reference, all speedups are reported
relative to it.
TVM MetaSchedule is the primary non-LLM baseline for kernel generation, which implements the same iterative feedback-loop, 
it updates XGBoost as its state to score mutated kernel variants to propose next candidates. We use default settings of TVM MetaSchedule for all the experiments.

\textbf{Agent pipelines.} The \textit{Sampling Agent} generates five candidates independently, converts
each outcome into structured feedback, and conditions a sixth candidate on all
five reference-candidate pairs.
The \textit{Feedback Loop} pairs an \emph{advisor\,/\,diagnoser} agents with a
\emph{coder} over up to five sequential iterations,  the advisor recommends
further optimizations when the kernel is correct, or the diagnoser prescribes a fix
when it is not. Pipeline diagrams are provided in
Appendix~\ref{sec:appendix-agent-diagrams}. The CUDA advisor/diagnoser prompts,
structured-feedback format, sampling post-processing prompt, and generation
settings are reported in Appendix~\ref{sec:appendix-agent-prompts-protocol}.

\textbf{Models.} The feedback-loop experiment uses
\texttt{gpt-oss-120b}, \texttt{Qwen3-Coder-Next}~(80B)\footnote{In contrast with function optimization section we replaced DeepSeek-V3.2 with Qwen3-Coder-Next because the former consistently failed to generate valid TVM~IR.}, and
\texttt{DeepSeek-V4-Pro} (1.6T). Non-code-generation calls are sampled at temperature~$0.7$ and code generation calls at~$0.1$.
For \texttt{gpt-oss-120b} we use medium reasoning effort.
All kernel generation experiments were conducted on a server with 8
NVIDIA A100-SXM4-80GB GPUs, CUDA~12.6, PyTorch~2.9.0, and an AMD EPYC~7513
32-Core CPU (2.0~TiB RAM).

\paragraph{Results for Iterative Kernel Optimization with Feedback.}\mbox{}

\begin{figure*}[!t]
\centering
\includegraphics[width=0.98\textwidth]{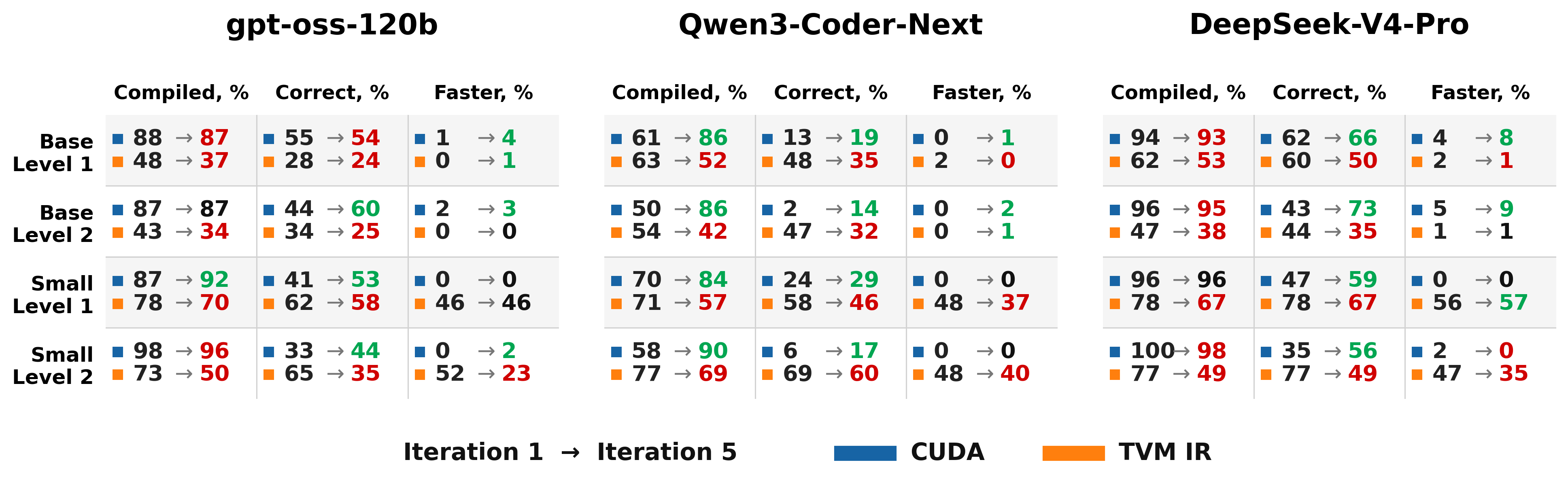}
\caption{Feedback-loop results for iterations 1 to 5. Colors at 5th iteration~$\mathbf{1\rightarrow \underline 5}$ denote  \textcolor{green!60!black}{\textbf{improvement}}, \textcolor{red}{\textbf{degradation}}, or  \textbf{no change}. Metrics are cumulative, not additive, \emph{faster} denotes the fraction of kernels that are correct and faster than \texttt{torch.compile}. Kernel difficulty  levels~1--2 are original KernelBench~\citep{ouyang2025kernelbench}. \emph{Base} is the original KernelBench configuration, and \emph{Small} is a
configuration with reduced input sizes.  
Intermediate iterations are presented in 
Appendix~\ref{sec:appendix-feedback-iterations}.}
\label{fig:cuda-tvm-feedback-validity}
\end{figure*}

\textbf{(1) Feedback behaves differently across the CUDA and TVM~IR
settings.} Figure~\ref{fig:cuda-tvm-feedback-validity} shows that, from the
first to the fifth iteration, compiled-and-correct rates generally increase
for CUDA across all three models. TVM~IR behaves in the opposite way:
compiled-and-correct rates decrease, while changes in correct-and-faster
rates are mixed.
Iterative feedback effectively steers proposals when the model has a dense
prior over the target representation, but it destabilizes generation when that
prior is sparse. Importantly, TVM~IR starts from a semantically stronger
reference, namely a valid unscheduled program, yet its pass rate still declines.
This highlights that the effect is driven by prior density rather than reference
quality.

\begin{figure}[!b]
\centering
\includegraphics[width=0.8\linewidth]{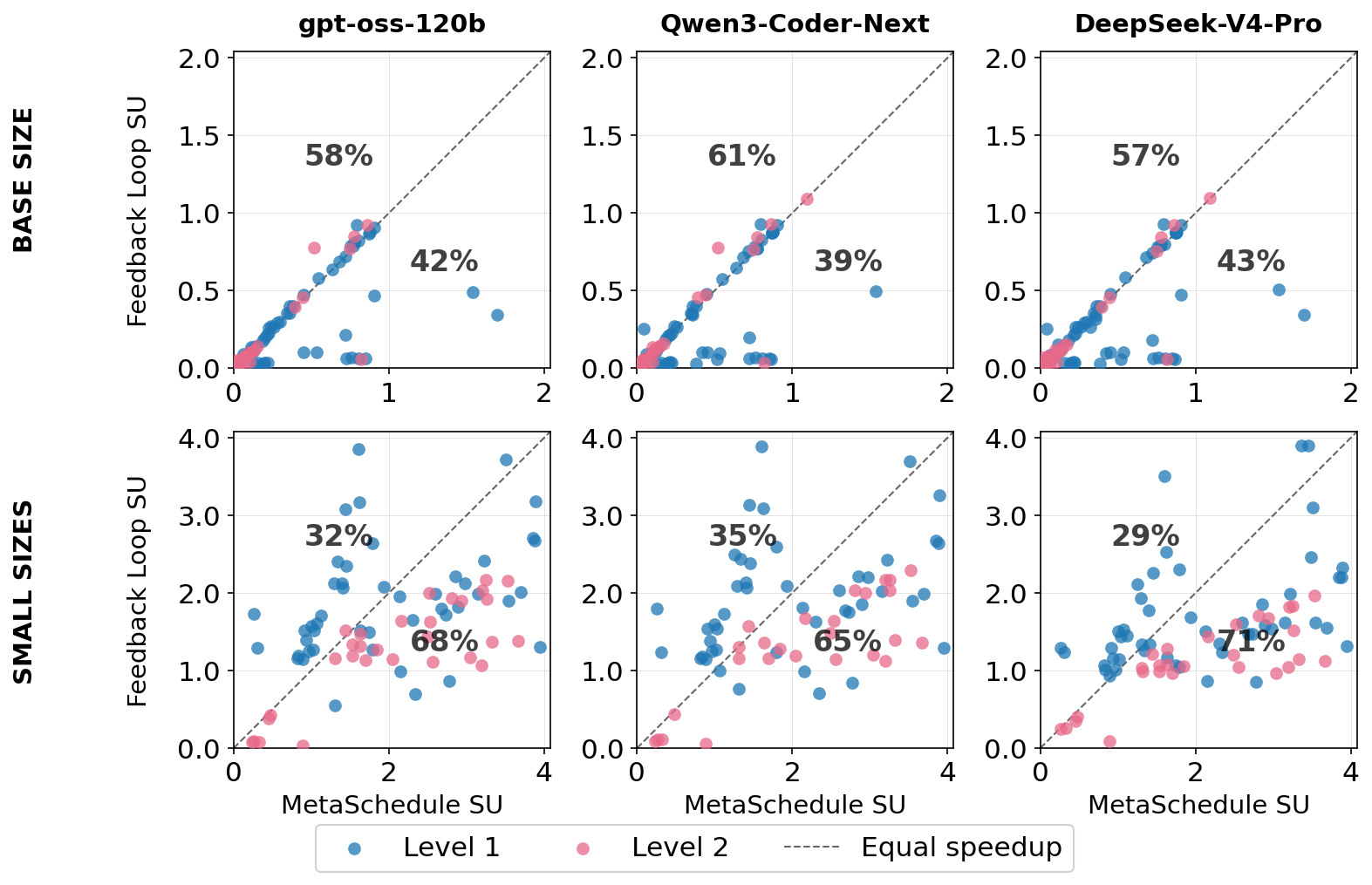}
\caption{Per-kernel speedup of TVM MetaSchedule against the
LLM-driven TVM~IR Feedback Loop. Points above the diagonal are kernels
where the LLM-generated kernels are faster, and percentages show the fraction of kernels in
each triangle. When input sizes change, MetaSchedule wins for most kernels and models.}
\label{fig:tvm-meta-feedback-scatter}
\end{figure}

\textbf{(2) Input size as a distribution shift.}
Figure~\ref{fig:tvm-meta-feedback-scatter} shows that in the \emph{Base}
regime, the LLM-driven TVM~IR feedback loop is roughly competitive with MetaSchedule.
In the \emph{Small} regime, the pattern reverses, MetaSchedule wins on $65$--$71\%$ of kernels, depending on the model,
with the effect strongest on Level~2 kernels. We hypothesize that small, atypical shapes are harder for the LLM because its
proposals carry parameter biases calibrated to
more common shapes, whereas MetaSchedule has no such prior and explores the  space freely.

\begin{figure*}[ht!]
\centering
\includegraphics[width=0.94\textwidth]{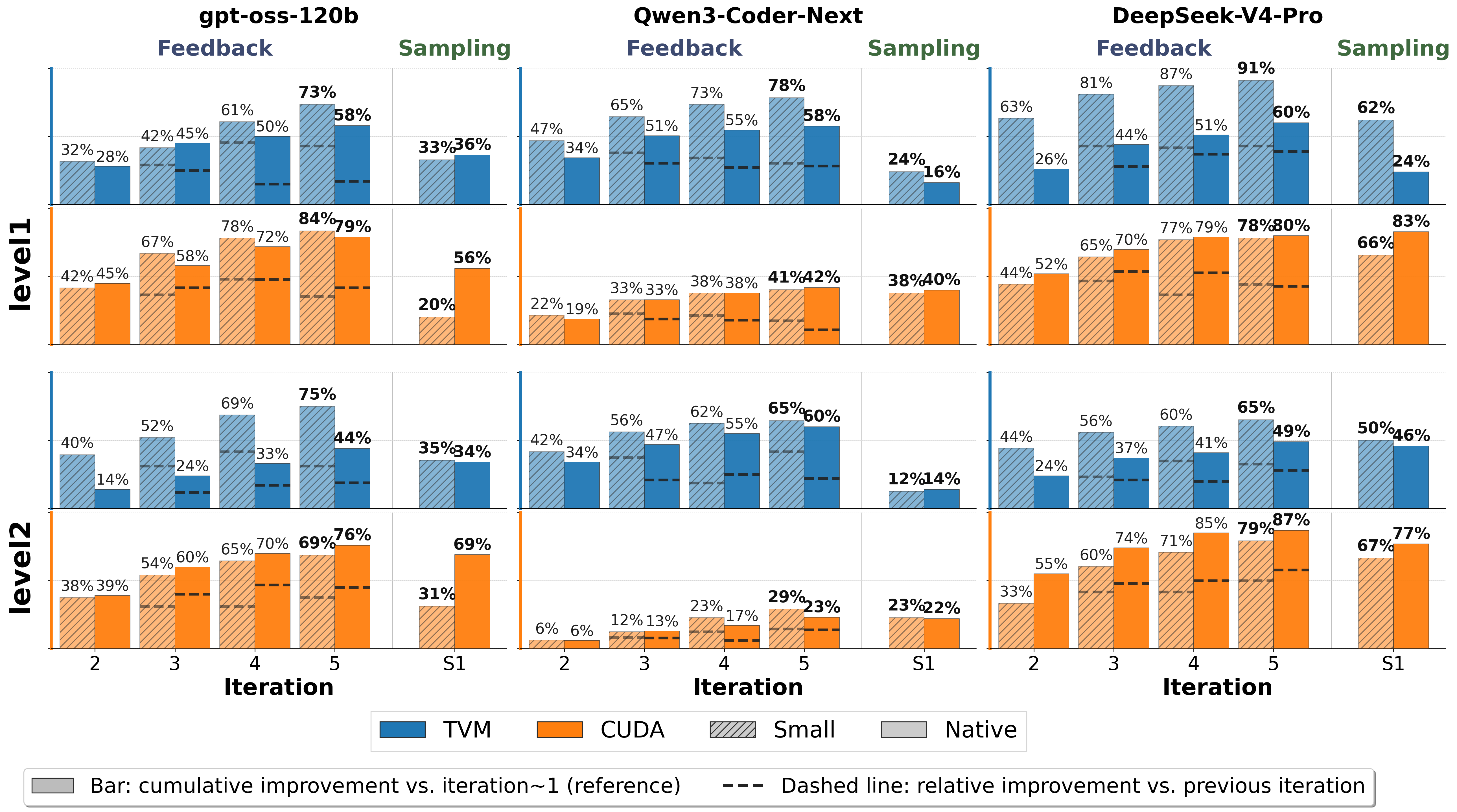}
\caption{Iterative improvement relative to the first iteration under two agent
structures (feedback loop vs.\ sampling ).
Feedback bars at iterations 2-5 show the cumulative fraction of kernels
improved over iteration~1. Dashed segments show per-step improvement.
The Sampling column (S1) shows the single post-aggregation candidate.}
\label{fig:feedback-vs-sampling-improvement}
\end{figure*}

\textbf{(3) Sequential feedback is better than parallel sampling.}
Figure~\ref{fig:feedback-vs-sampling-improvement} shows iterative gain 
relative to the iteration-1 baseline. Feedback loop based agent consistently 
outperforms the sampling agent at matched budget. The two converge 
only with CUDA in the \emph{Base} regime - densest-prior corner, elsewhere the gap 
is substantial. Two factors likely explain sampling's underperformance, an inflated context window and the fact that many small iterative corrections are better in general rather than one big update.

Relative gains are greater for TVM~IR, but only because CUDA starts from a stronger LLM-generated baseline with less headroom to improve. This does not contradict the 
absolute degradation in Figure~\ref{fig:cuda-tvm-feedback-validity}, 
since the improved and degraded kernel subsets are asymmetric.

\begin{figure*}[ht!]
\centering
\includegraphics[width=0.88\textwidth]{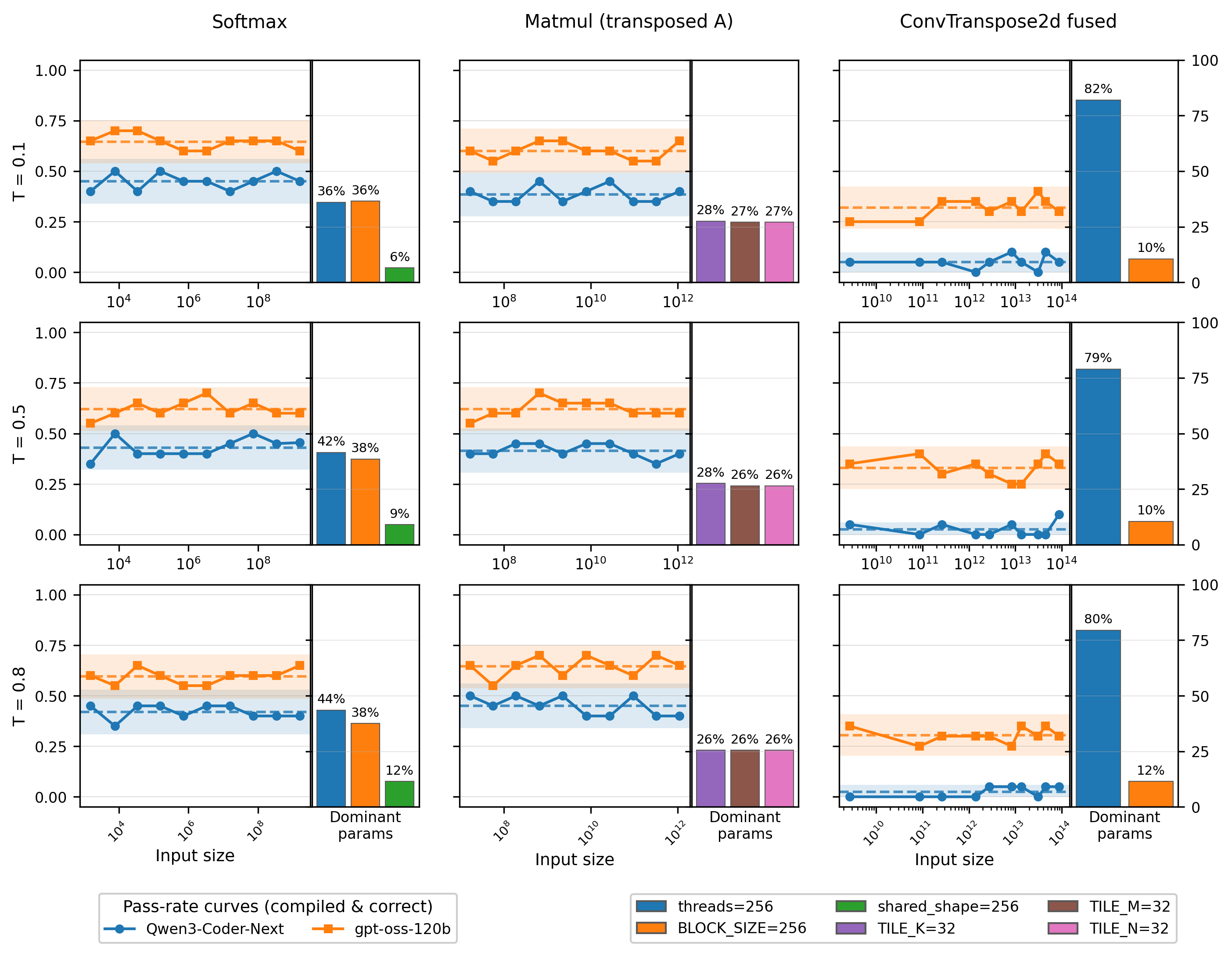}
\caption{Pass rate (left) and dominant kernel parameters (right) across shape grids
for three kernels, evaluated at different temperatures.
Pass rate is averaged over the
explicit and implicit prompt conditions, as both yield indistinguishable
curves. Flat pass rate means that generated kernels produce correct results regardless required input shapes. Parameter values are not constrained in the prompts, yet the models repeatedly generate the same values. For Matmul, three distinct values cover up to $78\%$ of generated kernels.}
\label{fig:zero-shot-pass-rate}
\end{figure*}

\subsection{Shape Bias in Kernel Generation}
\label{sec:zero-shot-prior}
Here we ask whether the model conditions on the requested input size at
all. To test this, we prompt models to generate CUDA kernels and specify the input shapes these kernels will be applied to. We construct a shape grid for three kernels~\footnote{(1) Softmax, (2) matrix multiplication with a transposed input, and (3) a convolution fused with batch normalization, Tanh, max pooling, and group normalization.}
spanning 10 input sizes from small to large.
Each prompt includes a Python reference implementation with concrete input
sizes.
We evaluate two instruction regimes: \emph{implicit} (sizes stated once in the
task description) and \emph{explicit} (the same, plus a direct instruction to
attend to the listed sizes).
We generate 20 kernels per grid point at different temperatures
$T\!\in\!\{0.1,0.5,0.8\}$, recording the pass rate and the dominant tiling
parameters in the generated kernels.
Full prompt templates are in Appendix~\ref{sec:appendix-shape-prompts}.

\textbf{Results for Shape-Bias Analysis.}
The left panel of Figure~\ref{fig:zero-shot-pass-rate} shows pass-rate curves
that are essentially flat across the entire size grid for both models~\footnote{\texttt{DeepSeek-V4-Pro} was not evaluated in this probe because the full shape grid was computationally too expensive for a 1.6T model.} and at
all three temperatures, and the explicit and implicit prompt conditions produce
nearly identical curves. The right panel shows the distribution of the parameters that control hardware
performance.\footnote{Block size, tile size, vector width, and unroll factor.} Although we did not restrict these parameter values, a small subset of values dominates the generated kernels, three distinct values cover up to $78\%$ of kernels. This collapse occurs regardless of input size or
temperature (see Appendix~\ref{sec:appendix-zero-shot-modes} for detailed results).
Together with the flat pass rate, this indicates that the models produce essentially the same kernel configurations regardless of input size.

One might argue that this is acceptable as long as the kernels remain correct across shapes. However, Appendix~\ref{sec:appendix-runtime-cost} shows that these generic parameters are not universally optimal. The best parameters depend on the input
size, and the model's fixed choices incur slowdowns of up to $1.76\times$.

We further ask whether kernel search is greedy, as in function optimization. To test this, we fix the kernel code and optimize only the parameter values. Appendix~\ref{sec:appendix-controlled-parameter-search} shows that LLMs explore a much narrower range of parameter values than Optuna-TPE, the same pattern we observed in black-box optimization (\S\ref{sec:setup-bbo}). Moreover, the parameters found by Optuna-TPE lead to faster kernels.

\begin{tcolorbox}[
    colback=white,
    colframe=black,
    boxrule=0.5pt,
    top=1mm, bottom=1mm
]
Across two experiments, LLM optimization is less reliable when the task is
formulated in TVM~IR or the input shape is atypical. Iterative feedback helps
in familiar domain - CUDA but hurts otherwise - TVM~IR.  MetaSchedule overtakes the LLM agent once froze prior is weak and is not updated.
Kernel generation reuses common parameter regardless tasked input sizes.
These observations are consistent with a frozen-weight-prior
interpretation in \S~\ref{sec:framework}.
\end{tcolorbox}

%% file: sections/06_discussion_conclusion.tex
\section{Discussion}
\label{sec:discussion}

\textbf{Prior alignment may shape context use.} Here, prior knowledge means information and behavioral biases encoded in frozen weights before the optimization episode, including effects of pretraining and post-training. Prompts, history, feedback, and search state instead belong to the context or pipeline (\S~\ref{sec:framework}). The greedy local search in BBO (\S~\ref{sec:llm-vs-optuna}) and the mode collapse in shape-conditioned kernel generation (\S~\ref{sec:results}) are consistent with the model favoring contextual information aligned with its frozen-weight prior. In BBO, the next decision anchors to the best current value. In kernel generation, the model reuses common parameter modes without mapping the requested shape into specialized tile, vector-width, or unrolling choices. We treat prior alignment as an empirical interpretation.


\textbf{LLM exploration can be improved in the following ways.}
(1) Either by adjusting prior via domain-specific RL~\citep{cudaagent2026}, (2) retrieving examples with the informative patterns at inference time and (3) by forcing exploration externally through agentic scaffolding.

\textbf{Limitations and Broader Impact.} Our study is limited to a few model families and standard benchmarks. The Stack counts are only a proxy for representation prevalence in public code. The evaluated models' actual pretraining and post-training mixtures are unknown, so Figure~\ref{fig:thestack-dist} does not directly measure their training exposure. Moreover, CUDA and TVM~IR differ in task formulation as well as representation, preventing a causal attribution to prior familiarity alone. Better kernel generation could reduce energy costs, but auto-generated code demands rigorous validation before production use.

\section{Conclusion}
\label{sec:conclusion}
This work argues that the central variable in agentic optimization is not
whether feedback exists, but whether feedback can be absorbed by the model's
prior.
Across BBO, shape-conditioning, and hardware optimization, we observe a common
pattern. LLMs are strong prior exploiters and weak open-ended searchers.
From this view, three takeaways follow.

First, feedback is representation-conditional. It is beneficial in dense-prior
domains and can destabilize in sparse-prior ones.
Second, sequential conditioning on concrete external signals is more effective
than one-shot aggregation of many sampled trajectories at similar budget.
Third, robust optimization pipelines should route adaptively. Use LLM
refinement where prior coverage is dense, and rely on task-adaptive search when
distribution shift or sparse prior makes language-only adaptation unreliable.
\clearpage

%% file: sections/99_appendix.tex
\section*{Appendix}
\label{sec:appendix}

The appendix is organized as follows.
Appendix~\ref{sec:related} surveys related work on LLM reasoning, LLM-based optimization, and LLM-based kernel generation.
\TODO{Appendix~\ref{sec:appendix-entropy} reports the token-level entropy experiment across code representations.}
\TODO{Appendix~\ref{app:functions-physical-construction} describes how the Functions and Physical task sets were constructed.}
Appendix~\ref{sec:appendix-bbo-prompts} lists the exact prompt templates used in the synthetic BBO comparison.
Appendix~\ref{app:metrics} defines the evaluation metrics for the pure BBO experiments.
\TODO{Appendix~\ref{app:bbob-budget-sensitivity} repeats the BBOB-5D comparison at larger evaluation budgets.}
Appendix~\ref{sec:appendix-greedy-gradient} details the experimental protocol for the greedy-gradient probe.
Appendix~\ref{sec:appendix-shape-prompts} provides the prompt templates for shape-conditioned zero-shot generation.
Appendix~\ref{sec:appendix-zero-shot-modes} tabulates the dominant zero-shot parameter modes across temperatures.
\TODO{Appendix~\ref{sec:appendix-runtime-cost} verifies that runtime-optimal kernel parameters depend on input size and that the LLM's repeated canonical settings are not universal defaults.}
\TODO{Appendix~\ref{sec:appendix-controlled-parameter-search} reports a separate controlled experiment comparing greedy parameter exploration by the LLM with Optuna-TPE on fixed CUDA kernels.}
Appendix~\ref{sec:appendix-tvm-tir-reference} describes the pipeline for constructing the unscheduled TVM-TIR reference module.
\TODO{Appendix~\ref{sec:appendix-tvm-prompts} reproduces the analysis, code, and summary prompts of the TVM agent.}
\TODO{Appendix~\ref{sec:appendix-agent-prompts-protocol} reproduces the CUDA advisor/diagnoser prompts, structured-feedback format, and sampling post-processing prompt, and reports generation settings.}
Appendix~\ref{sec:appendix-size-table} lists tensor shapes for five example tasks in the Base and Small regimes.
Appendix~\ref{sec:appendix-agent-diagrams} illustrates the two agentic pipeline architectures via flowcharts and a summary table.
Appendix~\ref{sec:appendix-feedback-iterations} expands the feedback-loop validity figure to all five iterations.
Appendix~\ref{sec:appendix-bbob-traces} provides per-task BBOB optimization trace plots.

\section{Related Work}
\label{sec:related}

We survey prior work across three threads directly relevant to our study, namely\\
(i)~the limits of LLM reasoning under distributional shift,\\
(ii)~LLMs as optimization operators and search primitives, and\\
(iii)~LLM-based approaches to GPU kernel generation.

\textbf{LLM Reasoning.} A growing body of work challenges the assumption that frontier LLMs possess robust, generalizable reasoning, converging instead on a picture of brittle pattern matching that degrades rapidly under distributional shift. \cite{stechly2024chain} provide an early and influential demonstration of this through careful experimentation in the Blocksworld planning domain. Varying both the generality of in-context examples and the complexity of test problems, they find that Chain of Thought (CoT) prompting yields meaningful gains only when prompts are narrowly tailored to the specific problem class and that even these gains collapse quickly as problem size grows beyond what was demonstrated. 
\cite{zhao2025chain} extend and formalize this critique through an explicit distributional lens. Introducing a controlled training environment, they demonstrate that CoT behavior reflects a structured bias learned from in-distribution data, whose effectiveness is fundamentally bounded by the degree of mismatch between training and test distributions.  \cite{sun2025omega}  evaluates out-of-distribution generalization along three axes exploratory, compositional, and transformative using programmatically generated problem templates with adjustable complexity. Evaluations of frontier models reveal sharp, monotonic accuracy degradation as complexity increases. Crucially, reinforcement learning fine-tuning on low-complexity data yields strong in-distribution gains but provides only limited and rapidly plateauing improvements out-of-distribution, with transformative generalization showing virtually no improvement at all. In the domain of competitive programming, \cite{zheng2025livecodebench} corroborate these findings using a contamination-free benchmark of 584 problems drawn from Codeforces, ICPC, and IOI, continuously updated and annotated by Olympiad medalists. Even the best reasoning models achieve only 53\% accuracy on medium-difficulty problems and 0\% on hard problems, with failures concentrated on observation-heavy problems that require novel insights rather than the recall of known techniques. Fine-grained error analysis further reveals that conceptual failures dominate over implementation bugs, and that tool augmentation accounts for a substantial share of performance gains that do not reflect native reasoning ability. 
\cite{gu2026illusion} question whether LLMs can stochastically explore environments by evaluating their ability to perform independent and sequential sampling of random numbers, finding that both approaches fail standard independence tests. However, the authors show that models can reliably \emph{convert} externally provided random seeds to target distributions suggesting that the failure lies not in distributional understanding but in autonomous random generation. This knowing-doing gap offers a partial explanation for why tool augmentation so consistently boosts LLM performance in agentic settings.

\textbf{LLM-Based Optimization.} A complementary line of research reframes LLMs from problem-solvers to search operators, algorithm designers, and formal modelers within optimization loops \citep{HUANG2024101663}. \cite{yang2024large} pioneered the use of LLMs as iterative optimizers via Optimization by PROmpting (OPRO), while subsequent work has expanded this to using LLMs for translating natural language directly into rigorous mathematical optimization models \citep{10.24963/ijcai.2025/1192}. When deployed as evolutionary search operators, trajectory analyses reveal that effective LLM optimizers diverge significantly from general reasoning behaviors. \cite{zhang2026makesllmgoodoptimizer} find that strong optimizers act as \emph{local refiners}, producing frequent incremental improvements while maintaining a localized search region, whereas high novelty often leads to semantic drift without fitness gains. This behavioral nuance is reflected in empirical benchmarks. \cite{ferreira2026llmsbeatclassicalhyperparameter} show that classical algorithms like CMA-ES consistently outperform pure LLM agents in fixed hyperparameter spaces, though hybrid approaches that share the classical optimizer's internal state (e.g., their proposed \textbf{Centaur} framework) achieve state-of-the-art results by letting the LLM apply domain knowledge to well-guided proposals. Architecturally, this fusion is formalized by \cite{10.1145/3638530.3654393}, who introduce the Evolution Transformer to distill evolutionary strategy updates into a causal attention mechanism, offering a structured alternative to purely text-based LLM search.

\textbf{LLM-Based Kernel Generation.} A recurring pattern is the pairing of LLMs with explicit exploration mechanisms. A notable production-oriented system is \textbf{KernelEvolve}~\citep{kernelevolve2025}, developed for generating optimized kernels across heterogeneous accelerators (NVIDIA, AMD, MTIA). It integrates tree search, agentic retrieval, knowledge base of hardware constraints, and a profiling feedback. However, the source code is proprietary, and the paper includes no ablation studies. Consequently, it remains unclear which components critically contribute to the reported gains. An open-source counterpart is KernelEvo, built on the GigaEvo evolutionary algorithm framework~\citep{khrulkov2025gigaevoopensourceoptimization}. It provides an implementation of LLM-guided evolutionary kernel generation. Yet it lacks a peer-reviewed paper, systematic evaluation and any ablation studies. \textbf{CUDA Agent}~\citep{cudaagent2026} develops a large-scale reinforcement learning system for CUDA kernel generation, using a synthetic dataset of fused operators, a structured agent toolset, and PPO with multi-stage warm-up. The system reports strong results on KernelBench. However, the work is limited to CUDA and lacks comparison with classical black-box optimizers (e.g., TVM MetaSchedule).
Broadly, three strategies dominate, namely\\
(i)~tree search,\\
(ii)~population-based evolutionary diversification, and\\
(iii)~reinforcement learning that adapts the model weights themselves rather than only the context.


\section{\TODO{Token-Level Entropy Across Code Representations}}
\label{sec:appendix-entropy}

\TODO{\textbf{What this measures.} Figure~\ref{fig:thestack-dist} counts public
code, which is an indirect estimate of how much of each representation a model
saw. Here we test whether the gap also appears in the models' own output
distributions during code generation. Higher token-level entropy means
probability is spread over more possible next tokens and therefore indicates
less confident generation.}

\TODO{\textbf{Setup.} We evaluate \texttt{gpt-oss-120b} and
\texttt{Qwen3-Coder-Next} on seven kernel operations, namely ReLU, Softmax,
Matmul, LayerNorm, Conv2d, cross-entropy and KL divergence. Each operation is
generated in six target representations, CUDA, OpenCL, HIP, TVM~IR, SYCL and
Triton. All conditions use the same zero-shot system instruction, which asks
the model to act as an expert GPU kernel engineer and return only code. The
user prompt specifies the operation and the target representation, and across
the six matched versions only the requested target representation changes.}

\begin{figure*}[!t]
\centering
\includegraphics[width=0.92\textwidth]{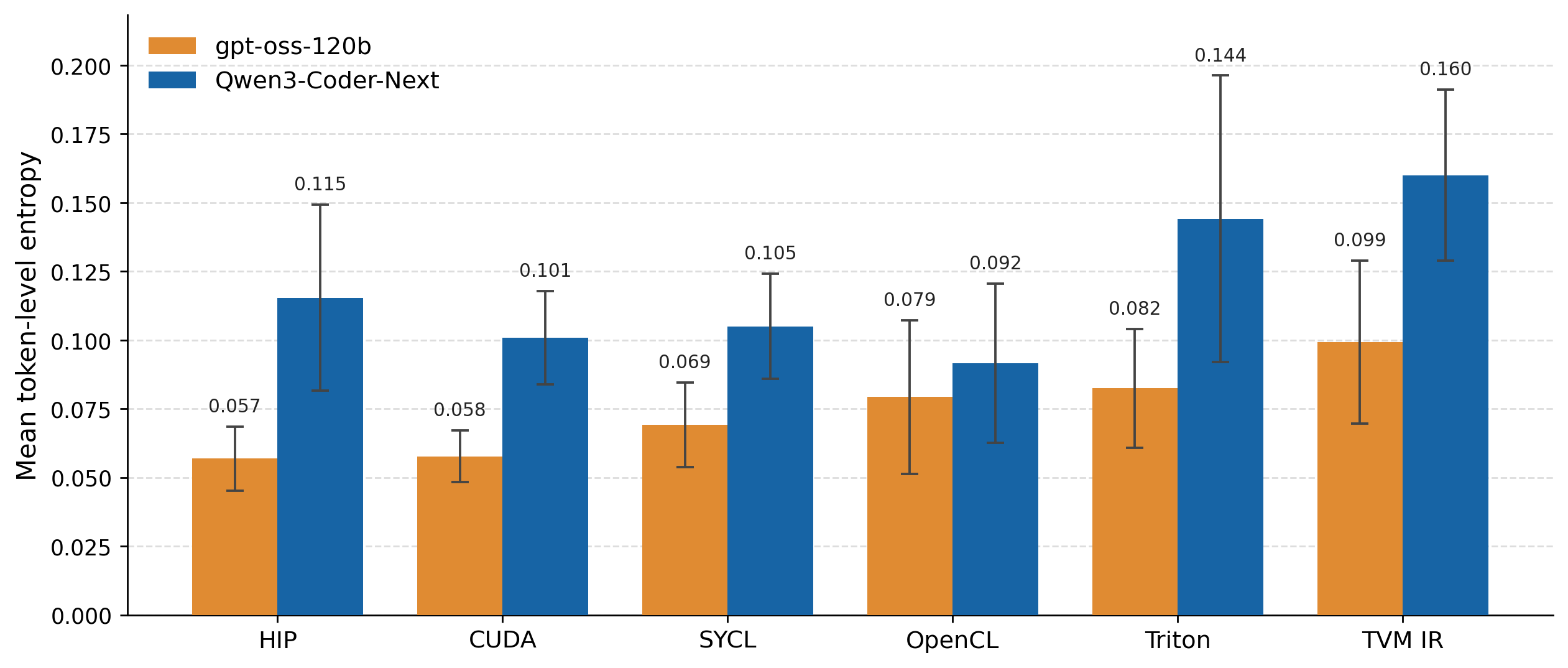}
\caption{\TODO{Mean token-level entropy per code representation, averaged over the
seven kernel operations of Table~\ref{tab:entropy-per-operation}. Bars are the operation means and error bars the
sample standard deviation across the seven operations. Orange is
\texttt{gpt-oss-120b} and blue is \texttt{Qwen3-Coder-Next}. Representations
are ordered by the \texttt{gpt-oss-120b} mean. TVM~IR is the highest of the six
representations for both models and CUDA is among the lowest.}}
\label{fig:entropy-by-representation}
\end{figure*}

\TODO{\textbf{Entropy computation.} We record the next-token distribution
produced at every generated code position. For vocabulary $\mathcal V$, the
entropy at position $t$ is}
\begin{equation}
\TODOcolor
h_t=-\sum_{v\in\mathcal V}p_t(v)\log p_t(v),
\label{eq:token-entropy}
\end{equation}
\TODO{where $p_t(v)$ is the probability assigned to token $v$. For an output
of $T$ code tokens, we compute the mean entropy
$\bar h=\frac{1}{T}\sum_{t=1}^{T}h_t$. Table~\ref{tab:entropy-per-operation}
reports this mean for every operation and
representation, and Figure~\ref{fig:entropy-by-representation} summarises each
representation as a mean and standard deviation over the seven operations.}

\begin{table*}[!t]
\centering
\caption{\TODO{Mean token-level entropy for every kernel operation and code
representation. TVM~IR is higher than CUDA for all seven operations under both
models.}}
\label{tab:entropy-per-operation}
\small
\setlength{\tabcolsep}{6pt}
\TODO{\begin{tabular}{llrrrrrr}
\toprule
\textbf{Model} & \textbf{Operation} & CUDA & OpenCL & HIP & TVM~IR & SYCL & Triton \\
\midrule
\texttt{gpt-oss-120b} & ReLU          & 0.044 & 0.048 & 0.055 & 0.121 & 0.062 & 0.066 \\
                      & Softmax       & 0.071 & 0.069 & 0.081 & 0.076 & 0.083 & 0.096 \\
                      & Matmul        & 0.049 & 0.073 & 0.047 & 0.067 & 0.054 & 0.069 \\
                      & LayerNorm     & 0.059 & 0.114 & 0.054 & 0.094 & 0.055 & 0.097 \\
                      & Conv2d        & 0.059 & 0.046 & 0.047 & 0.072 & 0.057 & 0.048 \\
                      & Cross-entropy & 0.067 & 0.112 & 0.054 & 0.121 & 0.087 & 0.093 \\
                      & KL divergence & 0.055 & 0.093 & 0.060 & 0.144 & 0.086 & 0.108 \\
\addlinespace[2pt]
\texttt{Qwen3-Coder-Next} & ReLU      & 0.112 & 0.090 & 0.099 & 0.147 & 0.104 & 0.129 \\
                      & Softmax       & 0.102 & 0.084 & 0.116 & 0.106 & 0.127 & 0.168 \\
                      & Matmul        & 0.064 & 0.059 & 0.064 & 0.206 & 0.117 & 0.083 \\
                      & LayerNorm     & 0.107 & 0.096 & 0.130 & 0.159 & 0.112 & 0.147 \\
                      & Conv2d        & 0.101 & 0.063 & 0.176 & 0.184 & 0.066 & 0.100 \\
                      & Cross-entropy & 0.105 & 0.103 & 0.115 & 0.156 & 0.106 & 0.243 \\
                      & KL divergence & 0.115 & 0.146 & 0.108 & 0.162 & 0.103 & 0.139 \\
\bottomrule
\end{tabular}}
\end{table*}

\FloatBarrier

\TODO{\textbf{Results.} TVM~IR has higher entropy than CUDA for every
evaluated operation under both models. Averaged across operations, entropy
increases from 0.058 for CUDA to 0.099 for TVM~IR with
\texttt{gpt-oss-120b}, a $1.71\times$ difference, and from 0.101 to 0.160 with
\texttt{Qwen3-Coder-Next}, a $1.58\times$ difference. TVM~IR also has the
highest mean among all six representations for both models, whereas CUDA is
among the lowest. The consistent CUDA--TVM~IR gap shows that both models
generate TVM~IR less confidently than CUDA under matched prompts. This
supports the interpretation that TVM~IR is a less familiar representation for
the evaluated models. The experiment establishes a generation-level
confidence difference, but does not by itself identify the source of that
difference.}

\section{\TODO{Construction of the Functions and Physical Task Sets}}
\label{app:functions-physical-construction}

\TODO{\textbf{Functions.} Each of the 100 tasks minimizes
$L(x,y)=|F(x,y)|$ over $(x,y)\in[0,1]^2$. With a fixed random seed, $F$ is
constructed by composing elementary shifted power and trigonometric terms.
Each expression combines polynomial components in $x$ and $y$, cross-variable
power terms, and sine or cosine components whose arguments depend on one or
both variables. Shifts are sampled uniformly from $[0.2,0.8]$, polynomial
degrees range from 2 to 5, trigonometric frequencies are integers from 4 to 9,
and all component weights are randomly sampled. A candidate expression is
retained only if it changes with both variables, is finite and nonconstant on
11 validation points.}

\TODO{\textbf{Physical tasks.} The 100 tasks are based on standard response
equations from five physical systems, with 20 instances per system, namely
$g(t;k,b)=kt+b$ for a spring, $g(t;k,b)=k\sin(bt)$ for a pendulum,
$g(t;k,b)=k\sin(2\pi bt)$ for alternating current,
$g(t;k,b)=k\exp(-bt)\cos(t)$ for a damped oscillator, and
$g(t;k,b)=k[1-\exp(-t/(b+0.01))]$ for an RC circuit. For each instance, hidden
parameters $(k^*,b^*)$ are sampled from the predefined search range and the
target trajectory is evaluated at 20 uniformly spaced time points
$t_j=j/19$. We add structured time-dependent noise generated from random
combinations of degree-1--3 polynomial and sine/cosine functions of time, with
the noise contribution scaled by 0.1. Thus, the target values are
$z_j=g(t_j;k^*,b^*)+0.1p(t_j;k^*,b^*)$. For a proposed $(k,b)$, the same
task-specific perturbation is used to obtain predictions
$\hat z_j=g(t_j;k,b)+0.1p(t_j;k,b)$, and the loss is the root-mean-square
error $L(k,b)=\sqrt{\frac{1}{20}\sum_{j=0}^{19}(\hat z_j-z_j)^2}$.}

\section{Prompt Templates for the Synthetic BBO Comparison}
\label{sec:appendix-bbo-prompts}

This section reports the exact prompt templates used when asking the LLM to act
as a black-box optimizer in the synthetic comparison from
Section~\ref{sec:llm-vs-optuna}.

\textbf{Two-dimensional function optimization.}
\begin{verbatim}
You are optimizing an unknown black-box function.
Goal: suggest x and y that minimize this function (smaller is better).
Search bounds: x and y must be in [0, 1].
Trial number: <trial_num>.
Current best loss: <best_loss>.
History of all previous attempts (step | x | y | loss):
<history_lines>
Important: do NOT repeat any previous (x, y) pair from history.
Always propose a new point that was not tried before.
Return ONLY JSON in this exact format: {"x": <float>, "y": <float>}.
\end{verbatim}

\textbf{Physical-system parameter optimization.}
\begin{verbatim}
You are optimizing physical system parameters.
Goal: suggest k and b that minimize loss (smaller is better).
Search bounds: k in [<k_min>, <k_max>], b in [<b_min>, <b_max>].
Trial number: <trial_num>.
Current best loss: <best_loss>.
History of all previous attempts (step | k | b | loss):
<history_lines>
Important: do NOT repeat any previous (k, b) pair from history.
Always propose new parameters that were not tried before.
Return ONLY JSON in this exact format: {"k": <float>, "b": <float>}.
\end{verbatim}

\textbf{BBOB function optimization.}
\begin{verbatim}
You are optimizing an unknown black-box function (BBOB benchmark).
Goal: suggest parameters that minimize this function.
Search bounds: <bounds_str>.
Trial number: <trial_num>.
Current best loss: <best_loss>.
History of all previous attempts (step | x0 | x1 | loss):
<history_lines>
Important: do NOT repeat any previous parameter combination.
Always propose new parameters.
Return ONLY JSON: <keys_desc>.
\end{verbatim}

In the two-dimensional BBOB runs used in the main experiments,
\texttt{<bounds\_str>} has the form
\texttt{x0 in [<lower\_0>, <upper\_0>], x1 in [<lower\_1>, <upper\_1>]},
and \texttt{<keys\_desc>} is
\texttt{\{"x0": <float>, "x1": <float>\}}.

\section{Evaluation Metrics for Pure BBO Optimization Problems}
\label{app:metrics}

Each task $\tau$ fixes a bounded feasible region
$\mathcal{X}_\tau \subset \mathbb{R}^d$ (e.g.\ $[0,1]^2$ for the function
family or the task-specific box for BBOB and parameter identification) that
all four optimization settings see and must sample from. For each run we
record the sequence of queried points $\{x_1, \ldots, x_K\} \subset
\mathcal{X}_\tau$ (with $K = 50$) and summarize behavior through three
quantities.

\emph{Best step} $\tau^\star = \arg\min_{k \in \{1,\ldots,K\}} y_k$ is the trial
index at which the best loss is first attained, averaged over tasks. It
measures how quickly an optimizer locks onto its final solution within the
budget.

\emph{Coverage} reports how broadly the domain has been probed. Let $\bar{x}_k$
be the centroid of the first $k$ queries and $r_k$ the distance from $\bar{x}_k$
to the farthest of them,
\begin{equation}
    \bar{x}_k \;=\; \frac{1}{k}\sum_{i=1}^{k} x_i,
    \qquad
    r_k \;=\; \max_{1 \leq i \leq k} \|x_i - \bar{x}_k\|_2.
    \label{eq:cov-radius}
\end{equation}
\TODO{For a $d$-dimensional search space, coverage is the volume of the
resulting bounding $d$-ball divided by the volume of the search domain. The
volume of a radius-$r_k$ ball in $\mathbb{R}^d$ is}
\begin{equation}
    \TODOcolor
    V_d(r_k)
    \;=\;
    \frac{\pi^{d/2}}{\Gamma\!\left(\frac{d}{2}+1\right)}r_k^d,
    \qquad
    \mathrm{Cov}_k
    \;=\;
    \frac{V_d(r_k)}{\operatorname{Vol}_d(\mathcal{X}_\tau)}.
    \label{eq:coverage}
\end{equation}
\TODO{For the two-dimensional experiments, the numerator is
$V_2(r_k)=\pi r_k^2$. For BBOB-5D, it is
$V_5(r_k)=\frac{8\pi^2}{15}r_k^5$. Reported as a percentage,
$\mathrm{Cov}_k$ indicates roughly how large a fraction of the search domain
the optimizer has probed by step~$k$.}

\emph{Normalized trajectory length} quantifies the straightness of the search
path,
\begin{equation}
    L
    \;=\;
    \frac{\|x_K - x_1\|_2}{\displaystyle\sum_{k=1}^{K-1} \|x_{k+1} - x_k\|_2}
    \;\in\; (0, 1].
    \label{eq:L}
\end{equation}
A perfectly linear trajectory attains $L = 1$. At the opposite extreme, an
isotropic random walk with iid zero-mean increments of size~$s$ has total path
length $(K{-}1)s$ and $\mathbb{E}\|x_K - x_1\|^2 = (K{-}1)\,s^2$, so its
typical displacement scales as $s\sqrt{K}$ while its path length scales as
$Ks$. The ratio therefore obeys $L \approx 1/\sqrt{K}$, giving
$L \approx 0.14$ at $K = 50$. High $L$ means the optimizer moves mostly in one direction and low $L$ means
it jumps around.

\section{\TODO{Budget Sensitivity on BBOB-5D}}
\label{app:bbob-budget-sensitivity}

\TODO{We extended the BBOB-5D evaluation from the 50-step budget used in the
main experiments to 75 and 100 steps. Tables~\ref{tab:bbob-budget-gptoss}
and~\ref{tab:bbob-budget-deepseek} report pairwise outcomes and final coverage.
Each pair is ``plain-LLM wins / baseline wins'' over the 48 tasks. Coverage is
reported as a percentage.}

\TODO{The qualitative conclusions are unchanged at the larger budgets. For
\texttt{gpt-oss-120b}, MCTS retains the highest coverage, increasing from
69.83\% at 50 steps to 73.58\% and 74.06\% at 75 and 100 steps, whereas LLM
coverage remains much lower at 11.76\%, 15.21\%, and 17.12\%. For
\texttt{DeepSeek-V3.2}, MCTS coverage increases from 97.88\% to 98.13\% and
98.35\%, while LLM coverage increases from 17.31\% to 19.31\% and 25.78\%.
At every budget, each baseline also retains its pairwise advantage over the
plain LLM. Thus, increasing the evaluation budget improves exploration but
does not reverse the relative BBOB-5D behavior observed at 50 steps.}

\begin{table}[t]
\centering
\caption{\TODO{BBOB-5D budget sensitivity for \texttt{gpt-oss-120b}. Pairwise
entries are plain-LLM wins / baseline wins. Coverage values are percentages.}}
\label{tab:bbob-budget-gptoss}
\resizebox{\linewidth}{!}{%
\begin{tabular}{@{}c|ccc|cccc@{}}
\toprule
\TODO{Budget} & \TODO{CMA-ES} & \TODO{Centaur} & \TODO{MCTS} &
\TODO{LLM Cov.} & \TODO{CMA-ES Cov.} & \TODO{Centaur Cov.} & \TODO{MCTS Cov.} \\
\midrule
\TODO{50}  & \TODO{11/37} & \TODO{11/37} & \TODO{19/29} & \TODO{11.76} & \TODO{41.08} & \TODO{48.47} & \TODO{69.83} \\
\TODO{75}  & \TODO{9/39}  & \TODO{10/38} & \TODO{16/32} & \TODO{15.21} & \TODO{58.12} & \TODO{58.50} & \TODO{73.58} \\
\TODO{100} & \TODO{9/39}  & \TODO{8/40}  & \TODO{18/30} & \TODO{17.12} & \TODO{66.21} & \TODO{64.41} & \TODO{74.06} \\
\bottomrule
\end{tabular}}
\end{table}

\begin{table}[t]
\centering
\caption{\TODO{BBOB-5D budget sensitivity for \texttt{DeepSeek-V3.2}. Pairwise
entries are plain-LLM wins / baseline wins. Coverage values are percentages.}}
\label{tab:bbob-budget-deepseek}
\resizebox{\linewidth}{!}{%
\begin{tabular}{@{}c|ccc|cccc@{}}
\toprule
\TODO{Budget} & \TODO{CMA-ES} & \TODO{Centaur} & \TODO{MCTS} &
\TODO{LLM Cov.} & \TODO{CMA-ES Cov.} & \TODO{Centaur Cov.} & \TODO{MCTS Cov.} \\
\midrule
\TODO{50}  & \TODO{12/36} & \TODO{9/39} & \TODO{23/25} & \TODO{17.31} & \TODO{41.08} & \TODO{51.81} & \TODO{97.88} \\
\TODO{75}  & \TODO{8/40}  & \TODO{4/44} & \TODO{21/27} & \TODO{19.31} & \TODO{58.12} & \TODO{62.63} & \TODO{98.13} \\
\TODO{100} & \TODO{5/43}  & \TODO{2/46} & \TODO{19/29} & \TODO{25.78} & \TODO{66.21} & \TODO{69.45} & \TODO{98.35} \\
\bottomrule
\end{tabular}}
\end{table}

\FloatBarrier
\section{Experimental Protocol for the Greedy-Gradient Probe}
\label{sec:appendix-greedy-gradient}

This section describes the experimental protocol underlying the greedy-gradient visualization in Figure~\ref{fig:greedy-gradient-subplot}. We detail the sampling procedure, the prompt setup, and the two prompt regimes (Table~\ref{tab:bbo-prompt-regimes}) used to verify that local refinement behavior is robust to prompt phrasing. \TODO{Across all six model--regime combinations}, the model consistently anchors its proposals to the best observed point and steps toward the minimum in short local increments.

In each rollout, $10$ points are sampled uniformly in $[0,1]^2$ and evaluated.
The full history is then passed to the LLM, which proposes $5$ additional
points (steps~$11$-$15$). The best of the first $10$ points is highlighted in
red in Figure~\ref{fig:greedy-gradient-subplot} for visualization only. It is
indistinguishable from the other nine in the prompt the model sees.
The protocol is repeated for \TODO{three models} and two prompt regimes
(Table~\ref{tab:bbo-prompt-regimes}). \TODO{In all six conditions,} the geometry is
the same. The LLM implicitly identifies the lowest-loss point, anchors its
proposals to it, and steps toward the minimum in short local increments.
\TODO{The reasoning traces in Table~\ref{tab:greedy-gradient-quotes} below
confirm this. In every condition the model describes a gradient strategy to
justify a local perturbation of the best seen point.}

\begin{table}[htbp]
\centering
\scriptsize
\setlength{\tabcolsep}{5pt}
\renewcommand{\arraystretch}{1.15}
\caption{\TODO{Verbatim reasoning excerpts recorded after step~10 for all three
models and both prompt regimes. In every condition the model describes a
gradient strategy to justify a local perturbation of the best seen point.}}
\label{tab:greedy-gradient-quotes}
\begin{tabular}{@{}p{0.18\linewidth}p{0.38\linewidth}p{0.38\linewidth}@{}}
\toprule
\TODO{\textbf{Model}} & \TODO{\textbf{Black-box}} & \TODO{\textbf{BO-style prompt}} \\
\midrule
\TODO{\texttt{gpt-oss-120b}} &
\TODO{``Use some heuristic: maybe \textbf{gradient} descent approximated.
[\ldots] propose slight variations [around the best point].''} &
\TODO{``Could try \textbf{gradient}-like: maybe lower both slightly? [\ldots]
pick a promising point near best but not same.''} \\
\TODO{\texttt{DeepSeek-V3.2}} &
\TODO{``We can try to explore around minima. Use some local search: maybe
\textbf{gradient} descent approximated. [\ldots] try slight perturbations.''} &
\TODO{``Propose a point near best but slightly different to refine
\textbf{gradient}: maybe (0.57,0.46) or (0.59,0.48).''} \\
\TODO{\texttt{DeepSeek-V4-Pro}} &
\TODO{``A gradient-guided approach can be used to probe nearby regions and
adjust the parameters incrementally toward better solutions.''} &
\TODO{``Choose a nearby candidate with a small offset from the current best
point to obtain a more accurate estimate of the local gradient.''} \\
\bottomrule
\end{tabular}
\end{table}

\begin{table*}[htbp]
\centering
\scriptsize
\setlength{\tabcolsep}{5pt}
\renewcommand{\arraystretch}{1.1}
\begin{tabular}{@{}p{0.48\linewidth}p{0.48\linewidth}@{}}
\toprule
\textbf{Black-box prompt} & \textbf{BO-style prompt} \\
\midrule
\begin{minipage}[t]{\linewidth}
\begin{Verbatim}[fontsize=\scriptsize,breaklines=true,breakanywhere=true,breaksymbolleft={}]
You are optimizing an unknown black-box function.
Goal: suggest x and y that minimize this function (smaller is better).
Search bounds: x and y must be in [0, 1].
Trial number: <N>.
Current best loss: <loss>.
History of all previous attempts (step | x | y | loss):
<history>
Important: do NOT repeat any previous (x, y) pair from history.
Return ONLY JSON in this exact format: {"x": <float>, "y": <float>}.
\end{Verbatim}
\end{minipage}
&
\begin{minipage}[t]{\linewidth}
\begin{Verbatim}[fontsize=\scriptsize,breaklines=true,breakanywhere=true,breaksymbolleft={}]
You are optimizing an unknown black-box function.
Goal: suggest x and y that minimize this function (smaller is better).
Search bounds: x and y must be in [0, 1].
Trial number: <N>.
Current best loss: <loss>.
History of all previous attempts (step | x | y | loss):
<history>
Important: do NOT repeat any previous (x, y) pair from history.
Return ONLY JSON in this exact format: {"x": <float>, "y": <float>}.
Optimize like Bayesian optimization!
\end{Verbatim}
\end{minipage} \\
\bottomrule
\end{tabular}
\caption{The two prompt regimes used in
Figure~\ref{fig:greedy-gradient-subplot}. The \emph{BO-style} prompt appends
a single line. Everything else is identical.}
\label{tab:bbo-prompt-regimes}
\end{table*}

\FloatBarrier
\section{Prompt Templates for Shape-Conditioned Zero-Shot Generation}
\label{sec:appendix-shape-prompts}

\TODO{We use two closely matched prompt templates to test whether failures to
condition on input shape are caused by the location of the size information in
the prompt. Both templates state the concrete input dimensions exactly once,
inside the reference implementation in the \texttt{<reference>} block. In the
\emph{implicit} condition the prompt stops there. In the \emph{explicit}
condition the same dimensions are repeated at the end of the prompt as a
closing instruction that names them as the primary optimization target. The
implicit template can also omit the hypothesis requirement in single-shot mode
through \texttt{require\_hypothesis=not is\_single\_shot}.}

\textbf{Explicit target-size prompt.}

\begingroup
\begin{Verbatim}[fontsize=\scriptsize,breaklines=true,breakanywhere=true,breaksymbolleft={}]
Your task is to implement an optimized version of the kernel provided in the `<reference>` section.
Maximize the kernel's performance using various optimization strategies, ensure that the output remains numerically identical to the provided reference and avoid memory errors.

Strictly adhere to these requirements:

1. **Class Definition:** You must define a class named exactly `ModelNew`. This is the entry point used to instantiate your kernel.
2. **Inheritance:** The class must inherit from `torch.nn.Module`.
3. **Initialization:** `__init__(self, ...)` must accept the arguments provided by the reference implementation's `get_init_inputs()`.
4. **Forward Pass:** `forward(self, ...)` must accept the arguments provided by the reference implementation's `get_inputs()`.
   - *Example:* If the baseline `get_inputs()` returns `[x, y]`, your method signature must be `forward(self, x, y)`.
5. **Output:** The return value of `forward` must have the exact same shape and data type as the reference output.
6. Use torch.utils.cpp_extension.load_inline to compile C++/CUDA source strings.
7. Do not write any code except described above.
8. Make code agnostic to device number, allocate output on the same GPU as the input (e.g. using `input.options()`).
9. Start code with # Hypothesis: ... comment.

{% if hardware_info %}
**Target Hardware:**
{{ hardware_info }}{% if compute_capability %} (Compute Capability: {{ compute_capability }}){% endif %}
{% endif %}

<reference>
{{ reference_code }}
</reference>

Start the file with a single comment line: `# Hypothesis: <your plan>` - briefly describe which specific optimization technique you are applying and why you expect it to improve{% if last_code %} (fix compilation, fix correctness, or increase speedup){% else %} performance{% endif %}.

{% if example_pre and example_after and not last_code %}
Refer to the following example to see how to rewrite a PyTorch model to an Inline CUDA model.

<example_pre>
{{ example_pre }}
</example_pre>

<example_after>
# Hypothesis: ...
{{ example_after }}
</example_after>
{% endif %}

{% if history %}
**Previous attempts (oldest first):**
{% for entry in history %}
- Hypothesis: {{ entry.hypothesis or '(none)' }}
  Result: {% if not entry.result.compiled %}Compilation failed: {{ (entry.result.compilation_error or '') }}{% elif not entry.result.valid %}Incorrect: {{ (entry.result.correctness_info or '') }}{% if entry.result.speedup is not none %} (speedup: {{ "%.2f"|format(entry.result.speedup) }}x){% endif %}{% elif entry.result.speedup is none %}Correct but performance measurement failed or timed out{{ (': ' ~ entry.result.performance_info) if entry.result.performance_info }}{% else %}Correct, speedup: {{ "%.2f"|format(entry.result.speedup) }}x{% endif %}

{% endfor %}
{% endif %}

{% if last_code %}
**Last attempt:**
<last_attempt>
{{ last_code }}
</last_attempt>

**Result:**
{% if not last_result.compiled %}
Compilation failed:
{{ (last_result.compilation_error or '') }}
{% elif not last_result.valid %}
Incorrect results:
{{ (last_result.correctness_info or '') }}
{% if last_result.speedup is not none %}
Performance: {{ "%.2f"|format(last_result.speedup) }}x
{% endif %}
{% elif last_result.speedup is none %}
Correct but performance measurement failed or timed out{{ (': ' ~ last_result.performance_info) if last_result.performance_info }}
{% else %}
Correct, speedup: {{ "%.2f"|format(last_result.speedup) }}x
{% endif %}
{% endif %}

Now write the new version of the kernel.

**IMPORTANT:** Begin your response with `# Hypothesis: <your plan>`.

{% if target_dims_text %}
**IMPORTANT (target sizes):** Optimize and validate the kernel specifically for these dimensions: `{{ target_dims_text }}`.
Treat these sizes as the primary optimization target.
{% endif %}
\end{Verbatim}
\endgroup

\textbf{Implicit condition.}

\begingroup
\begin{Verbatim}[fontsize=\scriptsize,breaklines=true,breakanywhere=true,breaksymbolleft={}]
Your task is to implement an optimized version of the kernel provided in the `<reference>` section.
Maximize the kernel's performance using various optimization strategies, ensure that the output remains numerically identical to the provided reference and avoid memory errors.

Strictly adhere to these requirements:

1. **Class Definition:** You must define a class named exactly `ModelNew`. This is the entry point used to instantiate your kernel.
2. **Inheritance:** The class must inherit from `torch.nn.Module`.
3. **Initialization:** `__init__(self, ...)` must accept the arguments provided by the reference implementation's `get_init_inputs()`.
4. **Forward Pass:** `forward(self, ...)` must accept the arguments provided by the reference implementation's `get_inputs()`.
   - *Example:* If the baseline `get_inputs()` returns `[x, y]`, your method signature must be `forward(self, x, y)`.
5. **Output:** The return value of `forward` must have the exact same shape and data type as the reference output.
6. Use torch.utils.cpp_extension.load_inline to compile C++/CUDA source strings.
7. Do not write any code except described above.
8. Make code agnostic to device number, allocate output on the same GPU as the input (e.g. using `input.options()`).
{% if require_hypothesis %}9. Start code with # Hypothesis: ... comment.{% endif %}

{% if hardware_info %}
**Target Hardware:**
{{ hardware_info }}{% if compute_capability %} (Compute Capability: {{ compute_capability }}){% endif %}
{% endif %}

<reference>
{{ reference_code }}
</reference>

{% if require_hypothesis %}
Start the file with a single comment line: `# Hypothesis: <your plan>` - briefly describe which specific optimization technique you are applying and why you expect it to improve{% if last_code %} (fix compilation, fix correctness, or increase speedup){% else %} performance{% endif %}.
{% endif %}

{% if example_pre and example_after and not last_code %}
Refer to the following example to see how to rewrite a PyTorch model to an Inline CUDA model.

<example_pre>
{{ example_pre }}
</example_pre>

<example_after>
# Hypothesis: ...
{{ example_after }}
</example_after>
{% endif %}

{% if history %}
**Previous attempts (oldest first):**
{% for entry in history %}
- Hypothesis: {{ entry.hypothesis or '(none)' }}
  Result: {% if not entry.result.compiled %}Compilation failed: {{ (entry.result.compilation_error or '') }}{% elif not entry.result.valid %}Incorrect: {{ (entry.result.correctness_info or '') }}{% if entry.result.speedup is not none %} (speedup: {{ "%.2f"|format(entry.result.speedup) }}x){% endif %}{% elif entry.result.speedup is none %}Correct but performance measurement failed or timed out{{ (': ' ~ entry.result.performance_info) if entry.result.performance_info }}{% else %}Correct, speedup: {{ "%.2f"|format(entry.result.speedup) }}x{% endif %}

{% endfor %}
{% endif %}

{% if last_code %}
**Last attempt:**
<last_attempt>
{{ last_code }}
</last_attempt>

**Result:**
{% if not last_result.compiled %}
Compilation failed:
{{ (last_result.compilation_error or '') }}
{% elif not last_result.valid %}
Incorrect results:
{{ (last_result.correctness_info or '') }}
{% if last_result.speedup is not none %}
Performance: {{ "%.2f"|format(last_result.speedup) }}x
{% endif %}
{% elif last_result.speedup is none %}
Correct but performance measurement failed or timed out{{ (': ' ~ last_result.performance_info) if last_result.performance_info }}
{% else %}
Correct, speedup: {{ "%.2f"|format(last_result.speedup) }}x
{% endif %}
{% endif %}

Now write the new version of the kernel.

{% if require_hypothesis %}
**IMPORTANT:** Begin your response with `# Hypothesis: <your plan>`.
{% endif %}
\end{Verbatim}
\endgroup

\section{Dominant Zero-Shot Parameter Modes (Tabular Summary)}
\label{sec:appendix-zero-shot-modes}

The table below matches the bar charts in Figure~\ref{fig:zero-shot-pass-rate}
(main text). Each entry is the fraction of generations that fix the listed
literal hyperparameter value, aggregated over the full shape grid and averaged
\TODO{over the two prompt conditions of
Appendix~\ref{sec:appendix-shape-prompts}. Both state the input dimensions
inside the reference code, and the \emph{explicit} condition additionally
repeats them as a closing instruction, whereas the \emph{implicit} condition
does not.}  Both conditions yield identical mode frequencies,
so only the aggregate is shown.

The parameters are as follows. \texttt{threads} is the number of GPU threads per thread block.
\texttt{BLOCK\_SIZE} is the number of input elements processed by one block.
\texttt{shared\_shape} is the width of the shared-memory tile used in reductions.
\texttt{TILE\_K}, \texttt{TILE\_M}, \texttt{TILE\_N} are tile sizes along the
$K$, $M$, and $N$ loop-nest dimensions of the tiled matrix multiply.

\begin{table}[htbp]
\centering
\scriptsize
\setlength{\tabcolsep}{4pt}
\renewcommand{\arraystretch}{1.05}
\begin{tabular}{@{}p{0.23\linewidth}p{0.31\linewidth}ccc@{}}
\toprule
\textbf{Task} & \textbf{Parameter mode} & \textbf{$T=0.1$} & \textbf{$T=0.5$} & \textbf{$T=0.8$} \\
\midrule
Softmax & \texttt{threads=256} & 36.0\% & 41.5\% & 43.5\% \\
Softmax & \texttt{BLOCK\_SIZE=256} & 36.5\% & 38.5\% & 37.5\% \\
Softmax & \texttt{shared\_shape=256} & 6.5\% & 9.0\% & 11.5\% \\
\midrule
Matmul with $A^\top$ & \texttt{TILE\_K=32} & 27.5\% & 27.5\% & 25.5\% \\
Matmul with $A^\top$ & \texttt{TILE\_M=32} & 27.0\% & 26.5\% & 25.5\% \\
Matmul with $A^\top$ & \texttt{TILE\_N=32} & 27.0\% & 26.5\% & 25.5\% \\
\midrule
Fused conv kernel & \texttt{threads=256} & 82.0\% & 79.0\% & 79.5\% \\
Fused conv kernel & \texttt{BLOCK\_SIZE=256} & 10.5\% & 10.5\% & 11.5\% \\
\bottomrule
\end{tabular}
\caption{Dominant zero-shot parameter modes across temperatures.  Each value is
the fraction of generations using the listed parameter value, aggregated over
the full shape grid (explicit and implicit shape-prompt conditions combined).
The same few modes dominate at all sizes and temperatures, confirming that
generation is governed by a fixed prior rather than by size-conditioned
exploration.}
\label{tab:zero-shot-modes}
\end{table}

\section{\TODO{Shape-Dependent Runtime Cost of Canonical Parameters}}
\label{sec:appendix-runtime-cost}

\TODO{Figure~\ref{fig:zero-shot-pass-rate} shows that the model repeatedly emits
the same canonical parameters as the input dimensions change. This observation
has two possible interpretations. The parameters could be universal hardware
defaults, or the model could be failing to condition its choice on input size.
To distinguish these explanations, we introduce an ablation that directly
tests whether these parameters require shape-specific tuning. For
representative Softmax and Matmul kernels, we hold the implementation algorithm
fixed and manually sweep the \texttt{threads} and tile parameters for each
input shape. Table~\ref{tab:shape-runtime-ablation} compares the resulting
runtime-optimal configurations with the parameters most frequently emitted by
the LLM. The common settings, \texttt{threads=256} and
\texttt{TILE=32/32/32}, are competitive mainly for medium-sized inputs. For
smaller and larger inputs, the optimum often changes while the model continues
to emit the same values. The invariant parameters are therefore not universal
defaults. The model fails to use the supplied dimensions and incurs real
latency penalties of up to $1.76\times$ for Softmax and $1.68\times$ for
Matmul. This pattern is consistent with reliance on familiar configurations
associated with standard input regimes rather than shape-specific adaptation.}

\begin{table*}[htbp]
\centering
\caption{\TODO{Runtime-optimal parameters versus the most common parameters
emitted by the LLM in the zero-shot shape grid. Latencies are in milliseconds.
Slowdown is common-parameter latency divided by best latency.}}
\label{tab:shape-runtime-ablation}
\scriptsize
\setlength{\tabcolsep}{3.5pt}
\TODO{\resizebox{0.98\textwidth}{!}{%
\begin{tabular}{llllrrr}
\toprule
Operator & Input shape & Runtime-optimal parameter & Best latency & Common emitted parameter & Common latency & Slowdown \\
\midrule
Softmax & $32{\times}512$ & \texttt{threads=64} & \textbf{0.0064} & \texttt{threads=256} & 0.0065 & $1.01{\times}$ \\
Softmax & $256{\times}4096$ & \texttt{threads=256} & \textbf{0.0090} & \texttt{threads=256} & 0.0090 & $1.00{\times}$ \\
Softmax & $1024{\times}16384$ & \texttt{threads=1024} & \textbf{0.0948} & \texttt{threads=256} & 0.1671 & $1.76{\times}$ \\
\midrule
Matmul & $32{\times}32{\times}32$ & \texttt{TILE=16/8/64} & \textbf{0.0068} & \texttt{TILE=32/32/32} & 0.0070 & $1.03{\times}$ \\
Matmul & $32{\times}16{\times}32$ & \texttt{TILE=32/16/16} & \textbf{0.0068} & \texttt{TILE=32/32/32} & 0.0070 & $1.03{\times}$ \\
Matmul & $1024{\times}1024{\times}1024$ & \texttt{TILE=32/64/16} & \textbf{0.2772} & \texttt{TILE=32/32/32} & 0.4589 & $1.66{\times}$ \\
Matmul & $1024{\times}512{\times}1024$ & \texttt{TILE=32/32/8} & \textbf{0.1857} & \texttt{TILE=32/32/32} & 0.2457 & $1.32{\times}$ \\
Matmul & $4096{\times}4096{\times}4096$ & \texttt{TILE=64/64/64} & \textbf{16.1497} & \texttt{TILE=32/32/32} & 26.9230 & $1.67{\times}$ \\
Matmul & $4096{\times}2048{\times}4096$ & \texttt{TILE=64/64/64} & \textbf{8.0080} & \texttt{TILE=32/32/32} & 13.4761 & $1.68{\times}$ \\
\midrule
Fused ConvTranspose2d & $1{\times}4{\times}4{\times}8{\times}8$ & \texttt{threads=64} & \textbf{0.0077} & \texttt{threads=256} & 0.0079 & $1.03{\times}$ \\
Fused ConvTranspose2d & $1{\times}16{\times}32{\times}32{\times}32$ & \texttt{threads=256} & \textbf{0.0150} & \texttt{threads=256} & 0.0150 & $1.00{\times}$ \\
Fused ConvTranspose2d & $2{\times}32{\times}32{\times}32{\times}32$ & \texttt{threads=512} & \textbf{0.0357} & \texttt{threads=256} & 0.0361 & $1.01{\times}$ \\
\bottomrule
\end{tabular}%
}}
\end{table*}
\FloatBarrier

\clearpage
\section{\TODO{Greedy Exploration in Controlled CUDA Parameter Search}}
\label{sec:appendix-controlled-parameter-search}

\TODO{To test whether the greedy LLM behavior observed in black-box
optimization extends to kernel optimization, we compare the LLM with
Optuna-TPE in a controlled CUDA parameter-search setting. We use fixed CUDA
implementations for nine Matmul workloads with different matrix dimensions and
nine Softmax workloads with different vector lengths and batch sizes. The
\texttt{gpt-oss-120b} agent and Optuna-TPE search the same discrete grids of
launch and tiling parameters under a matched budget of 50 trials per workload.
Holding the kernel code and search space fixed removes generation validity as
a confound and allows us to compare both exploration breadth and the runtime
of the best configuration found.}

\begin{table}[htbp]
\centering
\caption{\TODO{Best kernel latency after 50 trials: Optuna-TPE versus
\texttt{gpt-oss-120b}. Wins count workloads with lower best latency;
averages are over the nine workloads in each group. Lower latency is better.}}
\label{tab:optuna-vs-llm-latency}
\small
\setlength{\tabcolsep}{5pt}
\TODO{\begin{tabular}{lrrrr}
\toprule
Group & Optuna wins & LLM wins & Avg Optuna (ms) & Avg LLM (ms) \\
\midrule
Matmul & 6 & 3 & 1.492 & 1.618 \\
Softmax & 8 & 1 & 0.026 & 0.032 \\
\midrule
Total & 14 & 4 & --- & --- \\
\bottomrule
\end{tabular}}
\end{table}

\TODO{Table~\ref{tab:optuna-vs-llm-latency} shows that Optuna-TPE also
finds faster kernel configurations under the same trial budget, winning on
14 of 18 workloads (6 of 9 Matmul and 8 of 9 Softmax workloads). Its average
best latency is lower in both groups: 1.492 versus 1.618 ms for Matmul and
0.026 versus 0.032 ms for Softmax. Thus, the difference is not limited to
exploration breadth: Optuna-TPE also achieves better final kernel runtime
on most evaluated workloads.}

\begin{figure*}[htbp]
\centering
\includegraphics[width=\textwidth]{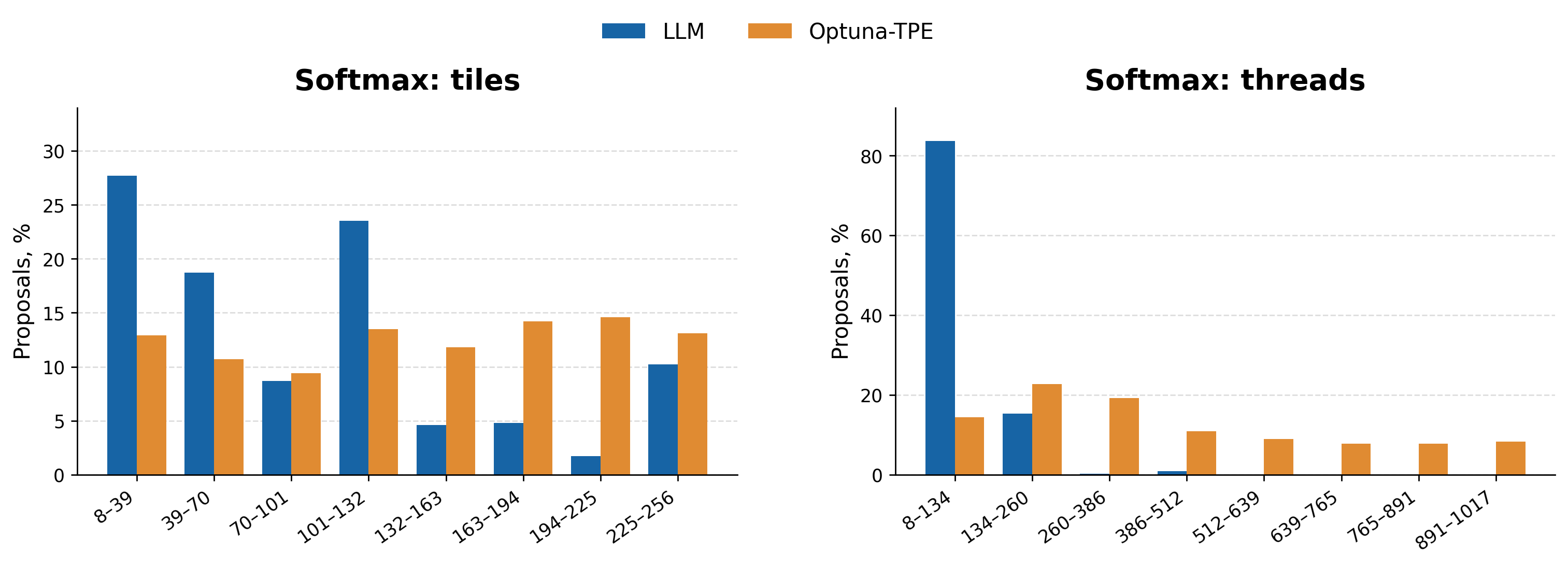}
\caption{\TODO{Softmax parameter proposals from the controlled CUDA search,
aggregated across nine workloads under a budget of 50 trials per workload.
Bars show the within-optimizer percentage of proposals in each parameter bin.
Blue denotes \texttt{gpt-oss-120b} and orange denotes Optuna-TPE. Concentration
in a narrow bin indicates repeated selection of the same parameter range.}}
\label{fig:controlled-search-softmax}
\end{figure*}

\begin{figure}[!t]
\centering
\includegraphics[width=\textwidth]{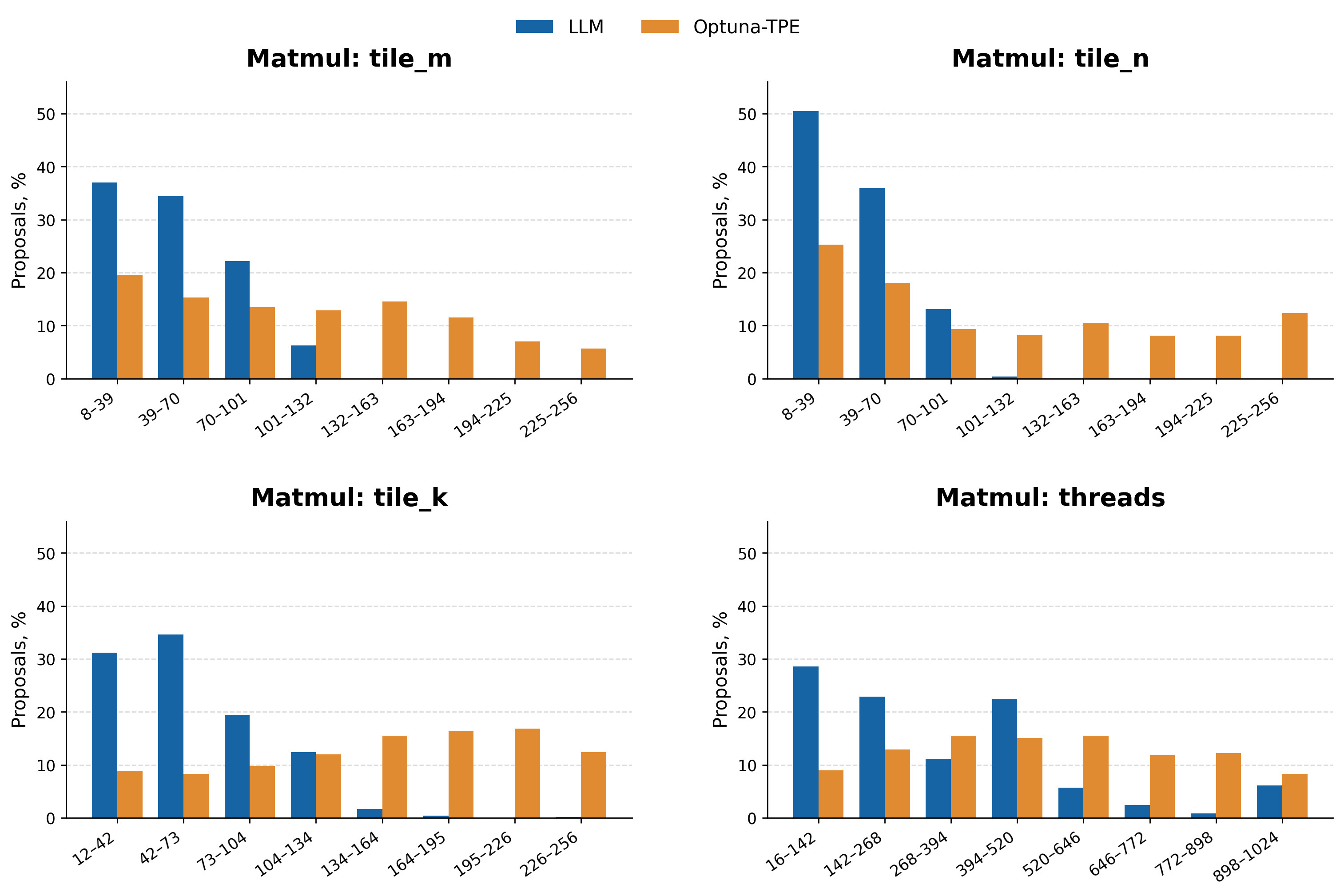}
\caption{\TODO{Matmul parameter proposals from the controlled CUDA search,
aggregated across nine workloads under a budget of 50 trials per workload.
Panels report \texttt{tile\_m}, \texttt{tile\_n}, \texttt{tile\_k}, and
\texttt{threads}. Bars show within-optimizer percentages for
\texttt{gpt-oss-120b} and Optuna-TPE. Concentration of the LLM mass in the
first few bins indicates repeated selection from a narrow parameter range.}}
\label{fig:controlled-search-matmul}
\end{figure}

\TODO{Figures~\ref{fig:controlled-search-softmax} and
\ref{fig:controlled-search-matmul} show the resulting proposal distributions.
The LLM concentrates in one or a few narrow bins, whereas TPE explores the
grid more broadly. For Softmax, 83.7\% of LLM \texttt{threads} proposals fall
in the lowest bin, compared with 14.4\% for TPE. For Matmul, 71.4\% of LLM
\texttt{tile\_m} proposals fall in the first two bins, and the LLM never
proposes a value above 132. Similar concentration appears for
\texttt{tile\_n} and \texttt{tile\_k}. This reproduces the greedy search
pattern from BBO in a controlled kernel-parameter task.}

\begin{table}[htbp]
\centering
\caption{\TODO{Standard deviation of parameter proposals in the controlled
CUDA search. Lower values indicate a narrower explored range.}}
\label{tab:controlled-search-std}
\setlength{\tabcolsep}{8pt}
\TODO{\begin{tabular}{llrr}
\toprule
Operator & Parameter & LLM std & Optuna-TPE std \\
\midrule
Softmax & \texttt{tiles} & 72.7 & 72.9 \\
Softmax & \texttt{threads} & 76.3 & 283.8 \\
\midrule
Matmul & \texttt{tile\_m} & 27.5 & 68.5 \\
Matmul & \texttt{tile\_n} & 22.0 & 78.9 \\
Matmul & \texttt{tile\_k} & 36.7 & 67.5 \\
Matmul & \texttt{threads} & 254.8 & 268.1 \\
\bottomrule
\end{tabular}}
\end{table}

\TODO{Table~\ref{tab:controlled-search-std} summarizes the same effect through
proposal variability. The LLM has substantially lower standard deviation for
Softmax \texttt{threads} and all three Matmul tile parameters. The difference
is negligible for Softmax \texttt{tiles} and modest for Matmul
\texttt{threads}, so the concentration is parameter-dependent rather than
universal. Together, the histograms and standard deviations show that the LLM
explores a narrower portion of the discrete kernel-parameter space than TPE.}

\section{Unscheduled TVM-TIR Reference Construction}
\label{sec:appendix-tvm-tir-reference}

Each TVM~IR experiment starts from an \emph{unscheduled} TensorIR module that
captures the computation graph of the KernelBench task without any loop
schedule, tiling, or parallelism annotations.  The module is produced by the
following fixed pipeline applied to the original PyTorch reference.

\textbf{Step 1. PyTorch $\to$ ONNX.}
The PyTorch \texttt{Model} class from each KernelBench task is traced and
exported to ONNX via \texttt{torch.onnx.export}.  When symbolic tracing fails (e.g.\ for models with
data-dependent control flow), the pipeline falls back to
\texttt{torch.fx.symbolic\_trace} and converts directly to TVM Relax IR.

\textbf{Step 2. ONNX $\to$ TVM Relax IR.}
The ONNX model is ingested by TVM's Relax ONNX frontend
(\texttt{tvm.relax.frontend.onnx.from\_onnx}) with all tensor shapes fixed
to the evaluation inputs of the task, producing a high-level Relax
\texttt{IRModule}.

\textbf{Step 3. Relax lowering to TIR.}
A deterministic sequence of seven Relax compiler passes is applied in order, namely
\texttt{LegalizeOps} (lower composite ops to TVM primitives),
\texttt{FoldConstant}, \texttt{CanonicalizeBindings}, \texttt{Normalize},
\texttt{FuseOps} (operator fusion),
\texttt{FuseTIR} (lower fused operator groups to \texttt{PrimFunc} blocks in
TensorIR), and \texttt{DeadCodeElimination}.
After \texttt{FuseTIR}, every computational kernel is represented as a
\texttt{T.prim\_func} with explicit loop nests over the output tensor axes,
but with no schedule annotations, so there is no loop splitting, no thread or
block binding, no vectorization, and no shared-memory staging.
The resulting module is what we call the \emph{unscheduled TVM-TIR reference}.

\textbf{What the LLM and MetaSchedule receive.}
Both the LLM agent and TVM MetaSchedule start from the same unscheduled
module.  MetaSchedule applies its tuning search within an explicitly
parameterised schedule space. The LLM agent receives the module serialised as
a \texttt{TVMScript} string (via \texttt{IRModule.script()}) and is asked to
return a rewritten version with schedule transformations applied.
This ensures that both baselines begin from the same computational
representation and that any performance difference is attributable solely to
the scheduling strategy, not to front-end conversion choices.

\TODO{\section{TVM-TIR Agent Prompts}\label{sec:appendix-tvm-prompts}}

\TODO{The TVM pipeline uses analysis, code, and summary agents. The analysis
and code agents receive a shared TIR reference with syntax examples and coding
constraints. Without this injected guidance, models rarely produced valid TVM
code, so we treat it as context rather than frozen-weight prior knowledge. Each
iteration also provides the current module, prior summaries and errors,
latency/profiling results, the best program, and analysis recommendations.}

\TODO{\subsection{Shared TIR/TVMScript syntax reference}}
\input{appendix_prompts/tir_syntax.tex}

\TODO{\subsection{Analysis-agent system prompt}}
\input{appendix_prompts/tir_analysis.tex}

\TODO{\subsection{Code-agent system prompt and output schema}}
\input{appendix_prompts/tir_code.tex}

\TODO{\subsection{Iteration-summary system prompt}}
\input{appendix_prompts/tir_summary.tex}

\TODO{\section{CUDA Agent Prompts and Reproducibility Settings}\label{sec:appendix-agent-prompts-protocol}}

\TODO{The CUDA pipelines use a coder together with an outcome-dependent
advisor or diagnoser. The advisor is invoked only after a correct candidate and
proposes a hardware-specific performance optimization. The diagnoser is invoked
after compilation or correctness failure and returns a three-line root-cause
analysis. The prompts below are the templates used by the implementation. Jinja
fields are retained to show exactly which reference code, candidate, history,
measurements, and profiler records are injected.}

\TODO{\subsection{Advisor prompt}}
\input{appendix_prompts/cuda_advisor.tex}

\TODO{\subsection{Diagnoser prompt}}
\input{appendix_prompts/cuda_diagnoser.tex}

\TODO{\subsection{Structured feedback format}}

\TODO{The shared feedback template records the reference, prior hypotheses and
outcomes, best-so-far candidate, latest candidate, timing statistics, and the
hottest profiled CUDA kernels. Outcomes are routed into four explicit cases,
compilation failure with compiler output, incorrect output with validation
details, correct output for which timing failed or timed out, and correct output
with mean latency, standard deviation, trial count, and profiler summary.}

\input{appendix_prompts/cuda_feedback_blocks.tex}

\TODO{For the Sampling Agent, each of the five candidate--environment pairs is
first converted into exactly one actionable sentence by the following
post-processing prompt. Candidate~6 receives the five code, environment-response,
and advice triples together. The Feedback Loop instead sends the corresponding
advisor or diagnoser output directly to the next coder call.}

\input{appendix_prompts/cuda_sample_advice.tex}

\TODO{The CUDA diagnoser uses temperature $0.7$ and at most 400 output tokens.
The advisor uses temperature $0.7$ and at most 512 output tokens. Coder calls
use temperature $0.1$, top-$p=0.95$, and at most 16,384 output tokens.}

\section{Benchmark Input-Size Variants}
\label{sec:appendix-size-table}

Table~\ref{tab:size-examples} lists the tensor shapes used in the Base
and Small regimes for five example tasks. Further details are available
in the accompanying repository. We use the following notation. $M$ and $N$ are the output
matrix dimensions (rows and columns), $K$ is the shared inner dimension of a matrix
product, \textit{batch} is the batch size, \textit{dim} is a feature or reduction
dimension, $C_{\mathrm{in}}$ and $C_{\mathrm{out}}$ are the input and output channel
counts of a convolution, $H$ and $W$ are spatial height and width, $k$ is the
convolution kernel size, and \textit{in}, \textit{hidden}, \textit{out} are the input,
hidden, and output feature dimensions of a linear layer.

\begin{table}[htbp]
\centering
\scriptsize
\setlength{\tabcolsep}{3pt}
\renewcommand{\arraystretch}{1.02}
\begin{tabular}{@{}p{0.06\linewidth}p{0.26\linewidth}p{0.28\linewidth}p{0.26\linewidth}@{}}
\toprule
\textbf{Level} & \textbf{Task} & \textbf{Base} & \textbf{Small} \\
\midrule
L1 & Standard matmul &
\begin{tabular}[t]{@{}l@{}}$M = 2048$\\$K = 8192$\\$N = 4096$\end{tabular} &
\begin{tabular}[t]{@{}l@{}}$M = 128$\\$K = 512$\\$N = 256$\end{tabular} \\[6pt]
L1 & Softmax &
\begin{tabular}[t]{@{}l@{}}batch $= 4096$\\dim $= 393216$\end{tabular} &
\begin{tabular}[t]{@{}l@{}}batch $= 4$\\dim $= 384$\end{tabular} \\[6pt]
L2 & \begin{tabular}[t]{@{}l@{}}Conv2D + ReLU\\+ bias add\end{tabular} &
\begin{tabular}[t]{@{}l@{}}batch $= 128$\\$C_{\mathrm{in}} = 64$\\$C_{\mathrm{out}} = 128$\\$H = W = 128$\\$k = 3$\end{tabular} &
\begin{tabular}[t]{@{}l@{}}batch $= 32$\\$C_{\mathrm{in}} = 16$\\$C_{\mathrm{out}} = 32$\\$H = W = 32$\\$k = 1$\end{tabular} \\[6pt]
L2 & \begin{tabular}[t]{@{}l@{}}Gemm + Sigmoid\\+ LSE\end{tabular} &
\begin{tabular}[t]{@{}l@{}}batch $= 16384$\\in $= 2048$\\hidden $= 4096$\\out $= 1024$\end{tabular} &
\begin{tabular}[t]{@{}l@{}}batch $= 512$\\in $= 64$\\hidden $= 128$\\out $= 32$\end{tabular} \\[6pt]
L2 & \begin{tabular}[t]{@{}l@{}}Gemm + Swish\\+ Divide + clamp stack\end{tabular} &
\begin{tabular}[t]{@{}l@{}}batch $= 1024$\\in $= 8192$\\out $= 8192$\end{tabular} &
\begin{tabular}[t]{@{}l@{}}batch $= 64$\\in $= 512$\\out $= 512$\end{tabular} \\
\bottomrule
\end{tabular}
\caption{Tensor shapes for five example tasks in the Base and Small
regimes. Values are taken directly from the matched
\texttt{level1}/\texttt{level1-small} and \texttt{level2}/\texttt{level2-small}
task definitions. Task labels are shortened for readability. Variable notation
is defined in the text above.}
\label{tab:size-examples}
\end{table}

\section{Agentic Pipeline Diagrams}
\label{sec:appendix-agent-diagrams}

Figure~\ref{fig:agent-diagrams} illustrates
the pipeline of each agentic architecture from two complementary perspectives,
a flowchart view and a graph-structure view.
Table~\ref{tab:agent-summary} summarizes the key properties of both architectures.

\begin{figure*}[htbp]
\centering
\resizebox{\textwidth}{!}{%
\begin{tikzpicture}[
  >=stealth, font=\scriptsize,
  box/.style={draw, rounded corners=3pt, fill=gray!8, text centered,
              minimum height=0.62cm, text width=2.9cm, inner sep=3pt},
  hibox/.style={draw, rounded corners=3pt, fill=blue!10, text centered,
                minimum height=0.62cm, text width=2.9cm, inner sep=3pt},
  arr/.style={->, thick}
]

\node[font=\small\bfseries] at (0, 0.5)  {(a) Sampling Agent};
\node[box]   (sa_ref)  at (0, 0)   {Reference code};
\node[box]   (sa_gen)  at (0,-1.3) {Generate 5 candidates\\(independent, same prompt)};
\node[box]   (sa_eval) at (0,-2.7) {Evaluate each\\compile \texttt{+} correctness \texttt{+} profiling};
\node[box]   (sa_fb)   at (0,-4.2) {Post-process\\error analysis or profiling\\summary ($5\times$)};
\node[hibox] (sa_out)  at (0,-5.5) {Generate candidate 6\\(aggregated feedback)};
\draw[arr] (sa_ref)  -- (sa_gen);
\draw[arr] (sa_gen)  -- (sa_eval);
\draw[arr] (sa_eval) -- (sa_fb);
\draw[arr] (sa_fb)   -- (sa_out);

\node[font=\small\bfseries] at (5.0, 0.5) {(b) Feedback Loop};
\node[box]   (fl_ref)  at (5.0, 0)   {Reference code};
\node[box]   (fl_adv)  at (5.0,-1.3) {Advisor / Diagnoser};
\node[box]   (fl_code) at (5.0,-2.7) {Coder};
\node[box]   (fl_eval) at (5.0,-4.2) {Evaluate\\compile \texttt{+} correctness \texttt{+} profiling};
\node[hibox] (fl_best) at (5.0,-5.5) {Best program};
\draw[arr] (fl_ref)  -- (fl_adv);
\draw[arr] (fl_adv)  -- (fl_code);
\draw[arr] (fl_code) -- (fl_eval);
\draw[arr] (fl_eval) -- (fl_best);
\draw[arr] (fl_eval.east) -- ++(0.8,0)
           -- ++(0, 2.9)
           node[midway, right, align=center] {$\leq$5\\iter.}
           -- (fl_adv.east);

\end{tikzpicture}
}
\caption{Pipeline diagrams for the two agentic architectures.
\textbf{(a)}~\emph{Sampling Agent}. Five candidates are generated independently
from the reference and evaluated. Post-processing converts each outcome into
structured feedback, which is aggregated to generate a sixth candidate.
\textbf{(b)}~\emph{Feedback Loop}. An advisor (on success) or diagnoser (on
failure) iteratively steers a coder stage for up to five iterations, with the
original reference and current best always in context.}
\label{fig:agent-diagrams}
\end{figure*}

\begin{table}[htbp]
\centering
\scriptsize
\setlength{\tabcolsep}{4pt}
\renewcommand{\arraystretch}{1.15}
\caption{Summary of the two agentic architectures used in
Section~\ref{sec:setup-kernel}. $\#$LLM is the maximum number of model calls
per task. The Sampling Agent uses 5 coder + 5 advisor/diagnoser + 1 coder and the
Feedback Loop uses 5 coder + 5 advisor/diagnoser.}
\begin{tabular}{@{}>{\raggedright\arraybackslash}p{0.17\linewidth}>{\raggedright\arraybackslash}p{0.17\linewidth}c>{\raggedright\arraybackslash}p{0.28\linewidth}>{\raggedright\arraybackslash}p{0.22\linewidth}@{}}
\toprule
\textbf{Agent} & \textbf{Representation} & \textbf{\#LLM} & \textbf{Search strategy} & \textbf{Feedback per step} \\
\midrule
Sampling Agent & CUDA, TVM~IR & 11 &
  Parallel sampling, 6th candidate conditioned on 5 aggregated feedbacks &
  Error analysis or profiling summary ($5\times$) \\
Feedback Loop & CUDA, TVM~IR & 10 &
  Sequential refinement, advisor or diagnoser selected by outcome &
  Advice, profiling, and best-so-far in context \\
\bottomrule
\end{tabular}
\label{tab:agent-summary}
\end{table}

\section{Full Feedback-Iteration Breakdown}
\label{sec:appendix-feedback-iterations}

\begin{figure*}[htbp]
\centering
\includegraphics[width=\textwidth,height=0.92\textheight,keepaspectratio]{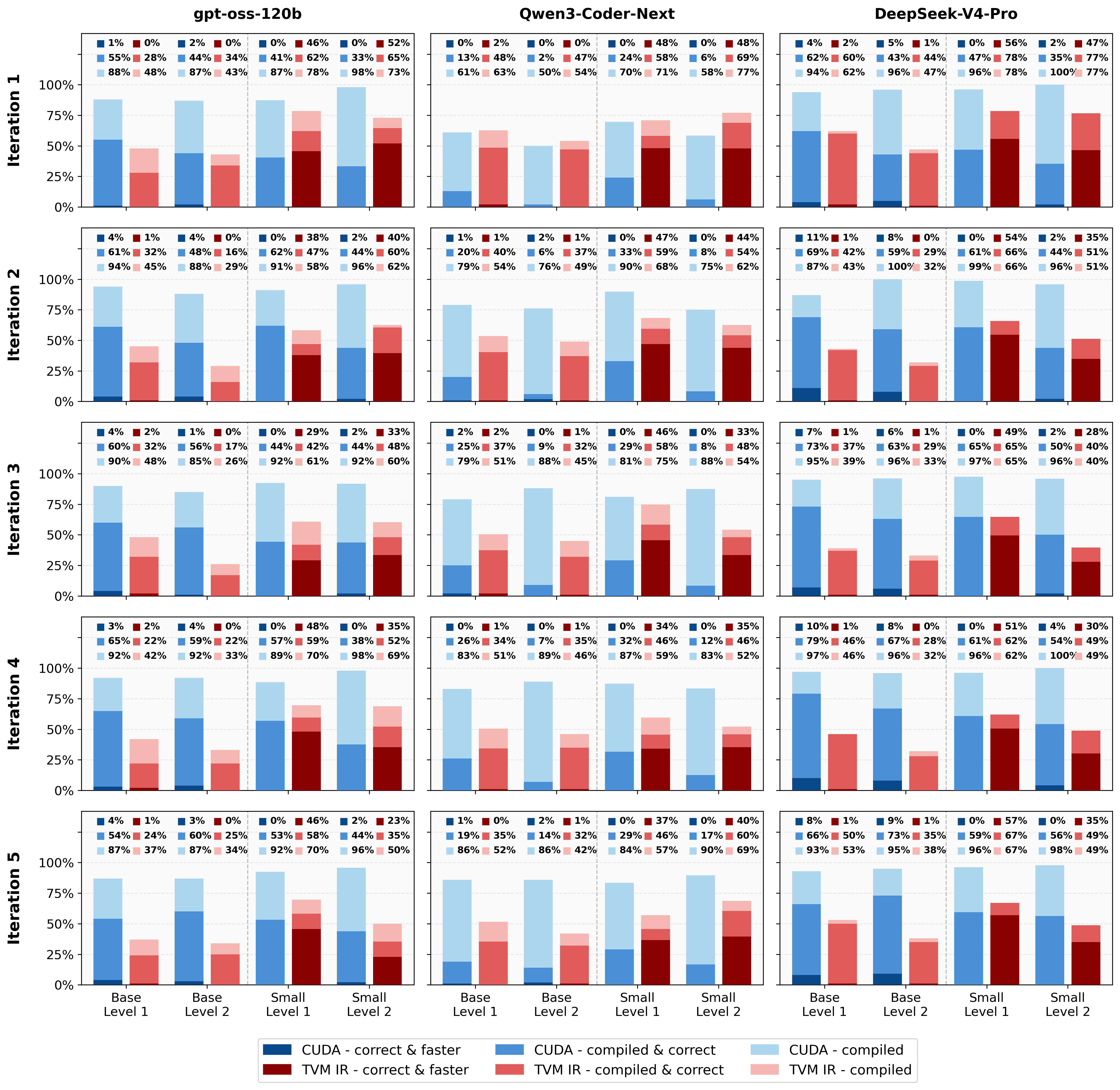}
\caption{Expanded Feedback Loop validity and speedup under CUDA and TVM~IR
generation across all five feedback iterations. Bars use the same nested
compile, correctness, and speedup encoding as the main-text figure.
\TODO{Panels cover \texttt{gpt-oss-120b}, \texttt{Qwen3-Coder-Next}, and
\texttt{DeepSeek-V4-Pro}.}}
\label{fig:cuda-tvm-feedback-all-iters}
\end{figure*}

Figure~\ref{fig:cuda-tvm-feedback-all-iters} expands
Figure~\ref{fig:cuda-tvm-feedback-validity} from the first and last feedback
iterations to the complete five-step trajectory.

\section{Additional BBOB Optimization Traces}
\label{sec:appendix-bbob-traces}

\begin{figure*}[t]
\centering
\includegraphics[width=\linewidth]{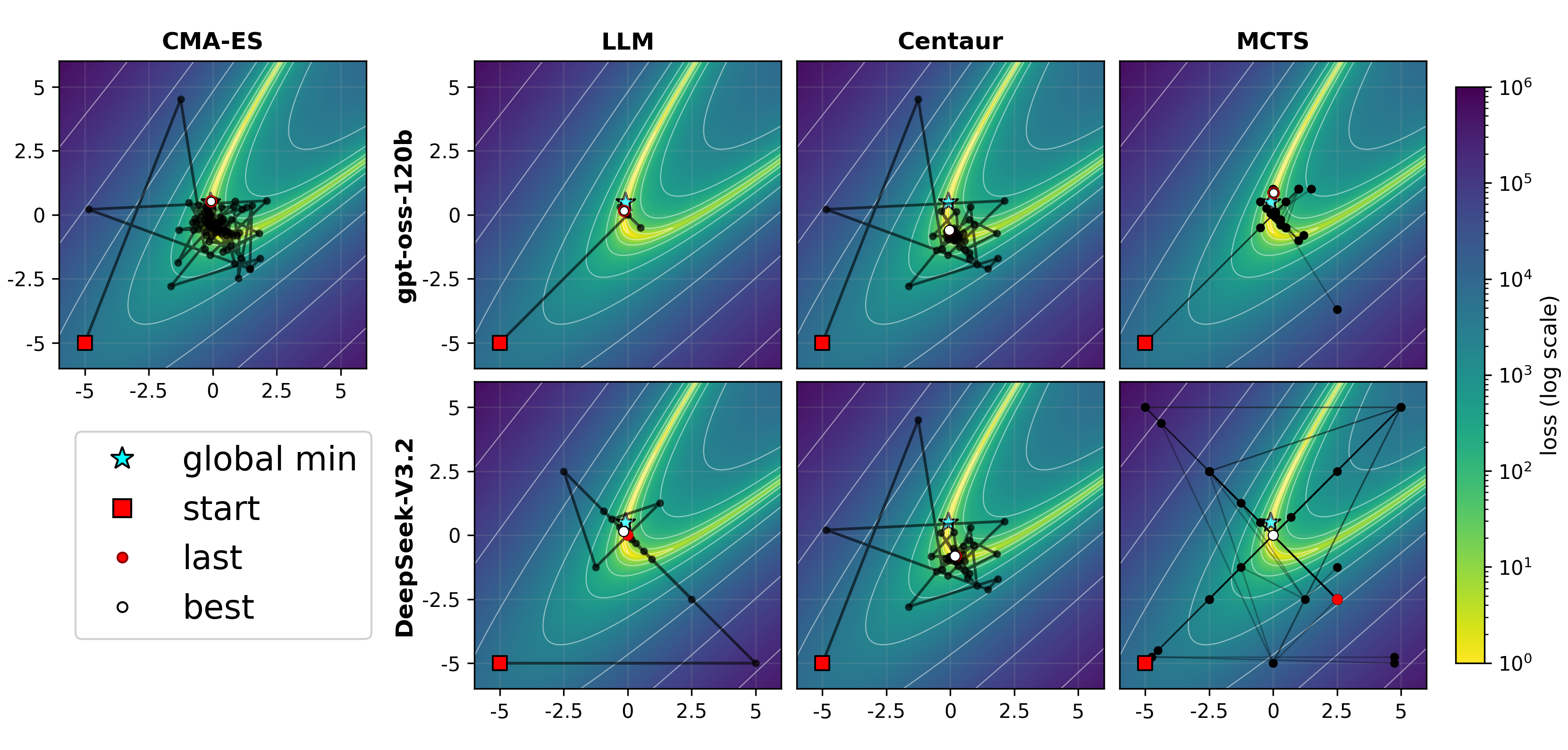}
\caption{BBOB task \texttt{bbob\_f09\_rosenbrock\_rotated\_i1}.}
\label{fig:bbobtrace-f09-rosenbrock-rotated-i1}
\end{figure*}

\begin{figure*}[t]
\centering
\includegraphics[width=\linewidth]{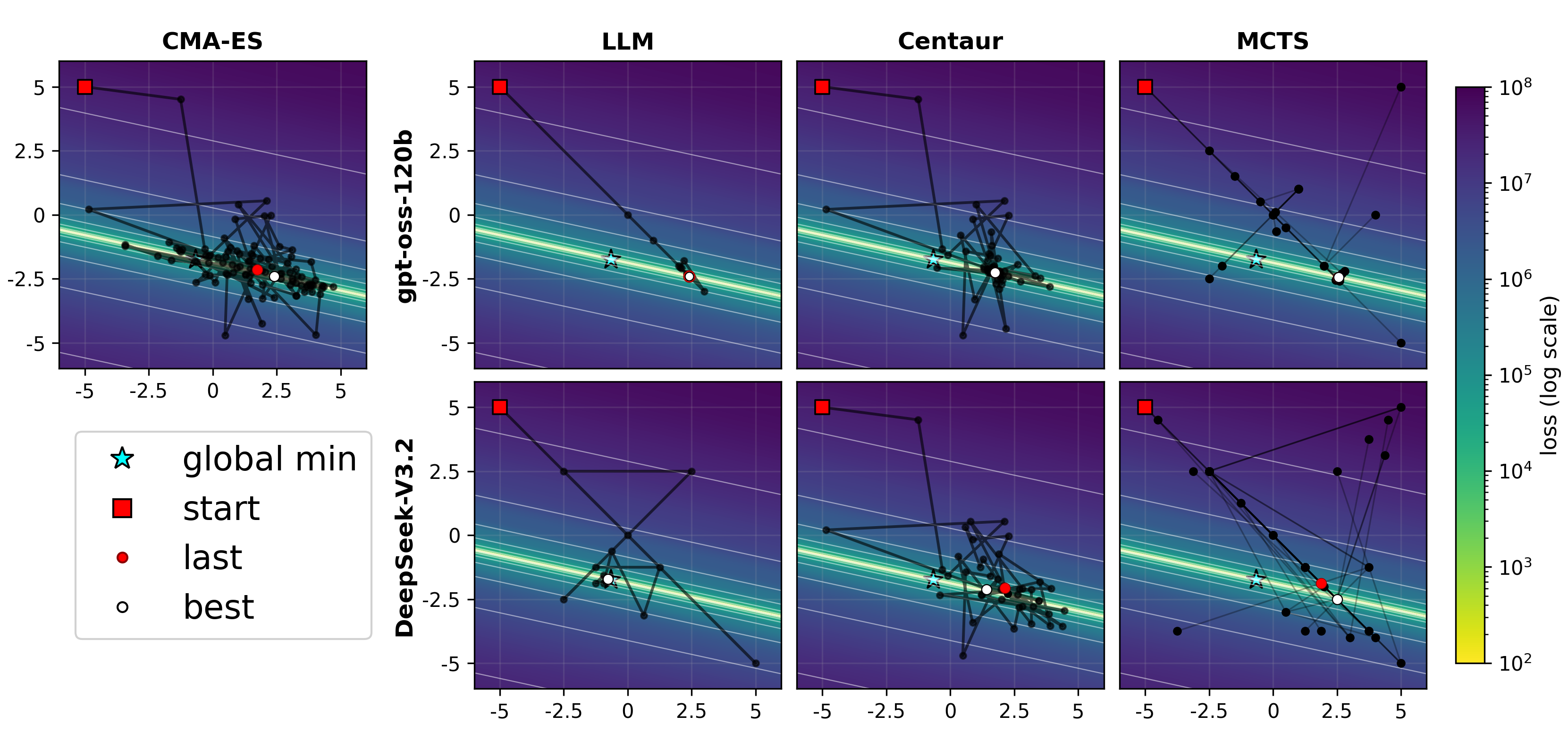}
\caption{BBOB task \texttt{bbob\_f10\_ellipsoid\_rotated\_i1}.}
\label{fig:bbobtrace-f10-ellipsoid-rotated-i1}
\end{figure*}

\begin{figure*}[t]
\centering
\includegraphics[width=\linewidth]{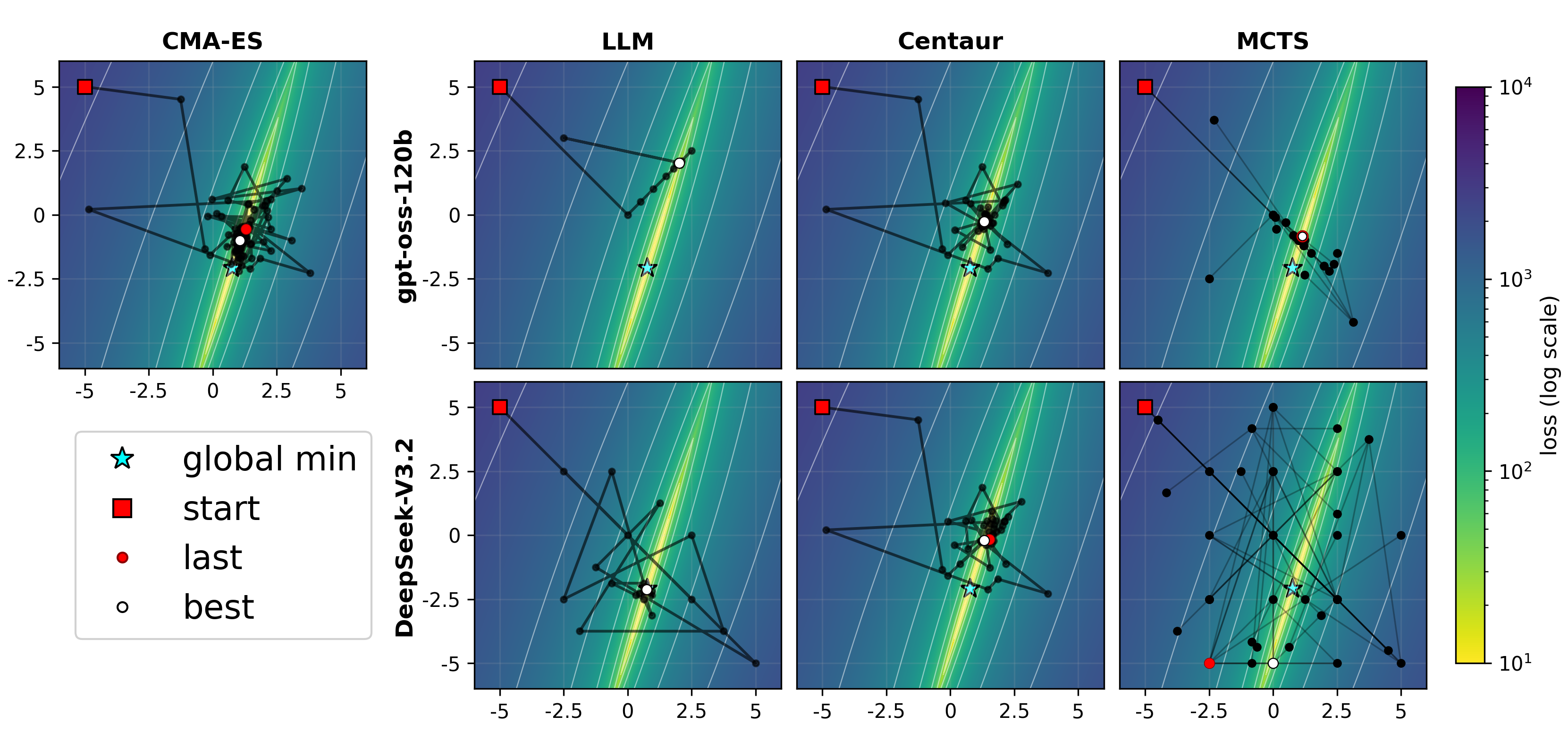}
\caption{BBOB task \texttt{bbob\_f13\_sharp\_ridge\_i1}.}
\label{fig:bbobtrace-f13-sharp-ridge-i1}
\end{figure*}

\begin{figure*}[t]
\centering
\includegraphics[width=\linewidth]{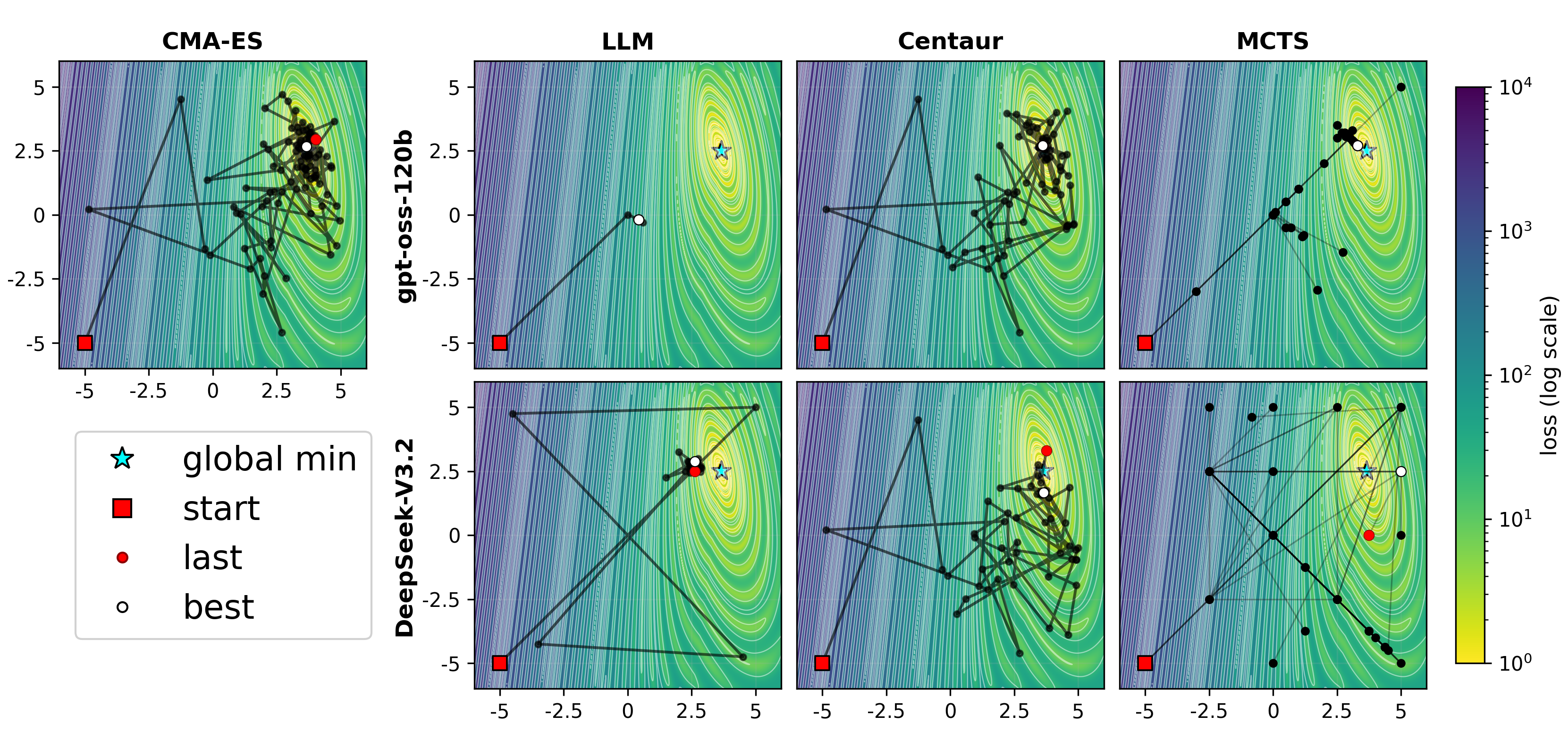}
\caption{BBOB task \texttt{bbob\_f17\_schaffers\_f7\_i1}.}
\label{fig:bbobtrace-f17-schaffers-f7-i1}
\end{figure*}

\begin{figure*}[t]
\centering
\includegraphics[width=\linewidth]{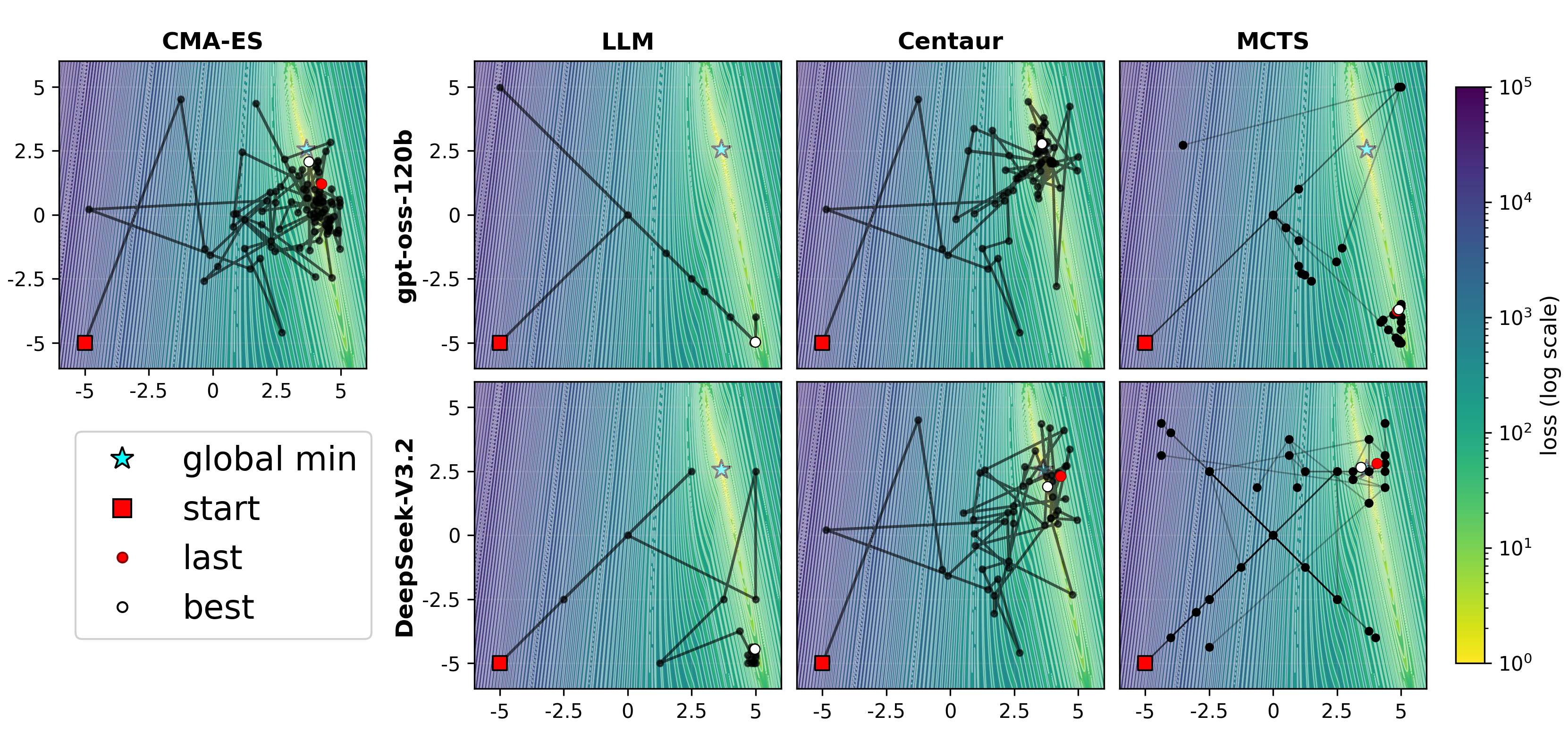}
\caption{BBOB task \texttt{bbob\_f18\_schaffers\_f7\_ill\_i1}.}
\label{fig:bbobtrace-f18-schaffers-f7-ill-i1}
\end{figure*}

\noindent
Figures~\ref{fig:bbobtrace-f09-rosenbrock-rotated-i1}-\ref{fig:bbobtrace-f18-schaffers-f7-ill-i1}
show additional BBOB optimization traces for
\texttt{gpt-oss-120b} and \texttt{DeepSeek-V3.2}.
Filenames match BBOB task identifiers
(\texttt{bbob\_fXX\_\ldots\_iY}).

%% file: appendix_prompts/tir_syntax.tex
\begin{Verbatim}[fontsize=\tiny,breaklines=true,breakanywhere=true]
# Base TIR/TVMScript syntax reference (included in tir_code and tir_analysis)
syntax: |
  ## TVM TIR / TVMScript syntax reference

  1. Imports and module structure
     - from tvm.script import ir as I, relax as R, tir as T
     - @I.ir_module class contains @R.function and @T.prim_func
     - Do NOT modify @R.function or R.call_tir; only optimize @T.prim_func bodies

  2. PrimFunc
     - @T.prim_func(private=True), def name(...) -> None
     - T.func_attr({"tir.noalias": True})
     - Return type -> None

  3. Buffers
     - T.Buffer((shape...), dtype): A: T.Buffer((128, 128), "float32")
     - T.match_buffer(handle, shape, dtype) for dynamic shapes
     - T.alloc_buffer((shape), dtype="float32", scope="global"|"shared"|"local")
     - Use scope= param, NOT .attach_scope()

  4. Loops
     - T.grid(n,m), T.serial(n), range(n)
     - T.parallel(n), T.unroll(n)
     - T.vectorized: LOOP only -- for v in T.vectorized(extent): or for v in T.vectorized(start, extent):
     - NEVER use with T.vectorized(): -- invalid; T.vectorized is a loop iterator, not a context manager
     - T.thread_binding(0, n, thread="blockIdx.x"|"threadIdx.x")
     - Thread tags: blockIdx.x/y/z, threadIdx.x/y/z

  5. Block (T.sblock)
     - with T.sblock("name"): ... defines computation block
     - Use T.sblock, NOT T.block
     - vi, vj, vk = T.axis.remap("SSR", [i, j, k]) -- S=spatial, R=reduce, O=opaque
     - T.reads(...), T.writes(...) -- use block vars (vi,vj), NOT loop vars (i,j)
     - with T.init(): for reduction blocks

  6. Expressions
     - T.float32(0), T.int64(), T.Cast("float32", x)
     - T.min, T.max, T.floor, T.abs, T.exp, T.sqrt, T.if_then_else

  7. CRITICAL -- T.axis.remap
     - MUST receive only loop iterator Var. NO constants, NO expressions, NO ForFrame.
     - WRONG: T.axis.remap("SS", [i0, T.int64(0)]) -> "Only Var is supported"
     - WRONG: T.axis.remap("SSR", [bx, tx, T.serial(8192)]) -> pass loop Var (k), NOT T.serial(...)
     - WRONG: T.axis.remap("SS", [bx, outer_tx*1024+tx]) -> expressions forbidden; use loop vars
     - WRONG: T.axis.remap("SS", [bx, k*1024]) -> expressions forbidden
     - WRONG: i0 = bx; T.axis.remap("SS", [i0, i1]) -> "Variable is not in the loop"
     - WRONG: T.axis.remap("SS", [v_ax0, v_ax1]) -> pass loop vars (i,j), NOT block vars (vi,vj) from remap.
     - String length must match bindings count: "SSR" -> 3 loop vars, "SSSSRR" -> 6 loop vars.
     - CORRECT: for i0 in T.thread_binding(...): for i1 in T.thread_binding(...): T.axis.remap("SS", [i0, i1])
     - Reduction axis (R) must be loop var: for k in T.serial(...): T.axis.remap("SSR", [i, j, k])
     - T.axis.remap, T.reads, T.writes MUST be inside with T.sblock("name"): -- else "The last frame is not SBlockFrame"

  8. CRITICAL -- R.function / R.call_tir (DO NOT MODIFY)
     - Do NOT modify @R.function body. Do NOT change R.call_tir arguments or out_sinfo.
     - metadata[...] is INVALID -- never use; R.call_tir receives inputs by name from R.function params.
     - Do NOT replace inputs with B, kernel, or invented names -- causes "Undefined variable: B", "Invoking function expects 2 arguments but 1 provided".
     - Do NOT use R.const(...).reshape(...) or other Relax ops -- "Constant object has no attribute reshape".
     - PrimFunc signature must match R.call_tir: do not add/remove params or change output shapes.
     - Do NOT change PrimFunc to return Tuple if R.call_tir expects single Tensor -- "out_sinfo must be compatible with PrimFunc".
     - Do NOT use cls.B or cls.Constant -- use the same arg names as in original R.call_tir; "Cannot find the function B".

  9. CRITICAL -- Do NOT use
     - metadata[...] -- invalid
     - T.block -- use T.sblock (T.block -> "module has no attribute 'block'")
     - .attach_scope() -- use scope= in T.alloc_buffer
     - with T.vectorized(): -- INVALID; T.vectorized is a loop: for v in T.vectorized(extent):
     - T.alloc_buffer outside T.sblock/prim_func -- "Block frame or PrimFunc frame not find"
     - T.alloc_buffer(..., scope="global") when targeting CUDA -- "Cannot allocate global memory"; pass as input
     - T.serial(a, b, c) -- only 1 or 2 args: T.serial(extent) or T.serial(start, extent)
     - Reassigning loop vars: remaining = remaining % ... -> "Variable remaining already defined"
     - T.texpr -- invalid; use T.Cast or numeric literals
     - T.block_attr -- invalid.
     - T.range -- invalid; use range(...) or T.serial(...).
     - T.Tuple -- invalid in TIR; PrimFunc returns single value.
     - for x in [a, b, c] -- invalid; loop must be range, T.serial, T.grid, etc., NOT Python list.
     - 2**step when step is Var -- "unsupported operand for **"; use constants or T.if_then_else, not Var in exponent.
     - min(1024, expr) in T.thread_binding -- use T.if_then_else or fixed extent; no Python min/and.
     - T.axis.reduce(extent) -- needs binding: T.axis.reduce(dom, loop_var); prefer T.axis.remap("R", [k]) with for k in T.serial(...).
     - T.axis.remap(...)[0] -- "'Var' object is not subscriptable"; use vi, vj = T.axis.remap("SS", [i, j]) directly.
     - R.call_tir(bare_name, ...) -- use cls.prim_func_name; "Undefined variable: cumprod_axis1".
     - "Cannot add loops on top of the root block" -- loops must wrap blocks; do not add orphan loops.
     - Loop var (tx, k) "defined outside of any block" -- for tx in T.thread_binding(...): must directly wrap the T.sblock that uses tx.
     - undefined tx/k/T_add/v_dh/pool_sum -- use only func params, loop iterators, block vars from remap; ensure loops enclose blocks.
     - "unexpected indent" -- emit valid Python/TVMScript indentation.

  10. GPU limits and buffer indexing
     - max_threads_per_block = 1024. threadIdx.x x threadIdx.y x threadIdx.z MUST <= 1024.
     - block=(8192,1,1) -> CUDALaunch CUDA_ERROR_INVALID_VALUE; split so threadIdx dim <= 1024.
     - For dim > 1024: split loop or flattened index.
     - Buffer indices count must match buffer rank: B[5D] needs 5 indices, lv[1D] needs 1; lv[()] for scalar is wrong for 1D.
     - "Cannot convert type float32x2048 to CUDA type" -- limit T.vectorized extent to 4, 8, 16, or 32; 2048 is too large.
     - v_ax2:int64 -> 0:int32 dtype mismatch: use T.int64(0) when loop var is int64, not raw 0.
     - "mismatched types int32 vs int64" in conditions: use T.Cast or match dtypes in comparisons.
     - block=(62,62,1): 62x62=3844 > 1024 -- threadIdx product MUST <= 1024.
     - "uses too much shared data" / "0x3fffc bytes, 0x29000 max" -- reduce shared buffer size.
     - "Duplicated block name" -- each T.sblock must have unique name per prim_func.
     - "Argument expected R.Tensor(...), given R.Tensor((), dtype=...)" -- don't pass scalar where tensor expected; preserve R.call_tir arg shapes.

  ---

  ## Example 1: Elementwise -- naive vs optimized

  Naive (T.grid):
    for i, j in T.grid(128, 128):
        with T.sblock("compute"):
            vi, vj = T.axis.remap("SS", [i, j])
            T.reads(A[vi, vj], B[vi, vj])
            T.writes(C[vi, vj])
            C[vi, vj] = A[vi, vj] + B[vi, vj]

  Optimized (thread binding):
    for i in T.thread_binding(0, 128, thread="blockIdx.x"):
        for j in T.thread_binding(0, 128, thread="threadIdx.x"):
            with T.sblock("compute"):
                vi, vj = T.axis.remap("SS", [i, j])
                T.reads(A[vi, vj], B[vi, vj])
                T.writes(C[vi, vj])
                C[vi, vj] = A[vi, vj] + B[vi, vj]

  ---

  ## Example 2: Matmul -- naive vs optimized (thread binding)

  Naive (T.grid):
    for i, j, k in T.grid(128, 128, 128):
        with T.sblock("C"):
            vi, vj, vk = T.axis.remap("SSR", [i, j, k])
            with T.init():
                C[vi, vj] = T.float32(0)
            T.reads(A[vi, vk], B[vk, vj])
            T.writes(C[vi, vj])
            C[vi, vj] = C[vi, vj] + A[vi, vk] * B[vk, vj]

  Optimized (thread binding on i,j; k stays reduction):
    for i in T.thread_binding(0, 128, thread="blockIdx.x"):
        for j in T.thread_binding(0, 128, thread="threadIdx.x"):
            for k in T.serial(128):
                with T.sblock("C"):
                    vi, vj, vk = T.axis.remap("SSR", [i, j, k])
                    with T.init():
                        C[vi, vj] = T.float32(0)
                    T.reads(A[vi, vk], B[vk, vj])
                    T.writes(C[vi, vj])
                    C[vi, vj] = C[vi, vj] + A[vi, vk] * B[vk, vj]

  For 128x128, blockIdx covers 128, threadIdx 128 -> 1 block. For larger: split i,j or use 2D grid.

  ---

  ## Example 3: Global sum reduction (HingeLoss-style) -- naive vs two-stage GPU

  Task: sum over BxN elements -> one scalar. Naive T.grid is slow.

  Naive:
    lv = T.alloc_buffer((B, N), "float32")
    for i, j in T.grid(B, N):
        with T.sblock("compute"):
            vi, vj = T.axis.remap("SS", [i, j])
            lv[vi, vj] = T.max(margin, T.float32(0.0))
    for i, j in T.grid(B, N):
        with T.sblock("sum"):
            vi, vj = T.axis.remap("RR", [i, j])
            with T.init():
                out[0] = T.float32(0.0)
            out[0] = out[0] + lv[vi, vj]
    out[0] = out[0] / T.float32(B * N)

  Optimized (two-stage: stage1 partials per block, stage2 reduce partials):
    stage1: sh = T.alloc_buffer((256,), "float32", scope="shared")
            for bx in T.thread_binding(4096, thread="blockIdx.x"):
                for tx in T.thread_binding(256, thread="threadIdx.x"):
                    sh[tx] = T.float32(0.0)
                    for it in T.serial(0, num_iters):
                        idx = bx * 256 + tx + it * (4096 * 256)
                        if idx < B * N:
                            sh[tx] = sh[tx] + ...
                    T.tvm_storage_sync("shared")
                    # tree reduction 256->128->64->32->16->8->4->2->1
                    if tx < 128: sh[tx] = sh[tx] + sh[tx + 128]
                    T.tvm_storage_sync("shared")
                    if tx < 64:  sh[tx] = sh[tx] + sh[tx + 64]
                    # ... down to 1
                    if tx == 0: partials[bx] = sh[0]
    stage2: one block reduces partials[0..4095] to scalar.

  Key: T.tvm_storage_sync("shared") before reading shared after writes.
\end{Verbatim}

%% file: appendix_prompts/tir_analysis.tex
\begin{Verbatim}[fontsize=\tiny,breaklines=true,breakanywhere=true]
# TIR Analysis agent: analyzes code + feedback, outputs plain text recommendations
system: |
  1. Top-level module structure

  1.1. IRModule
    - The root is an IRModule: @I.ir_module class containing @R.function and @T.prim_func.
    - Required imports (may be commented in script output):
        from tvm.script import ir as I
        from tvm.script import relax as R
        from tvm.script import tir as T
    - Do NOT modify @R.function bodies or R.call_tir references; only optimize @T.prim_func.

  2. TIR PrimFunc definition - Format, T.func_attr, T.Buffer, T.match_buffer, T.alloc_buffer.
  3. Loop constructs - T.grid, T.serial, T.thread_binding.
  4. Block structure - T.sblock, T.axis.remap, T.reads, T.writes, T.init().
  5. TIR expressions - literals, arithmetic, control flow.
  6. Memory - T.alloc_buffer scope global/shared/local.
  7. GPU scheduling - thread binding, shared memory, vectorization, tiling.
  8. Schedule transformations - split, tile, reorder, bind, vectorize.
  9. Reduction patterns - sum, max/min.
  10. Optimization guidelines - tiling, memory hierarchy, parallelism, vectorization.
  11. Common pitfalls -- error-specific fixes:
     - "Undefined variable: metadata" / "Undefined variable: B" / "expects 2 arguments but 1 provided": NEVER touch R.call_tir; only optimize T.prim_func.
     - "Only Var is supported in T.axis.remap": pass loop iterator vars, NOT T.int64(0), expressions (k*1024), or T.serial(...).
     - "Variable is not in the loop: ax0/v_ax0": use loop vars (i,j,k) in remap, NOT block vars (vi,vj) or computed vars (ax0=bx).
     - "Expected Array<ir.PrimExpr> but got ForFrame": pass loop Var (k), not for k in T.serial(...).
     - "kinds.size() == bindings.size()": remap string length must match loop count (e.g. "SSR" -> 3 vars).
     - "The last frame is not SBlockFrame": T.axis.remap, T.reads, T.writes must be inside with T.sblock("name"):.
     - "Block frame or PrimFunc frame not find" for T.alloc_buffer: put T.alloc_buffer inside with T.sblock(...) or at prim_func top.
     - "CUDALaunch CUDA_ERROR_INVALID_VALUE" / block=(8192,1,1): threadIdx.x (or product) MUST <= 1024; split loop.
     - "Buffer X is N-dimensional, cannot be indexed with M-dimensional indices": indices count = buffer rank.
     - T.block -> use T.sblock. T.serial(3 args) -> use T.serial(extent) or T.serial(start, extent).
     - "Variable remaining already defined": do not reassign loop vars (e.g. remaining = remaining % n) inside loop.
     - "Constant object has no attribute reshape": do not use R.const(...).reshape() in R.call_tir; only modify T.prim_func.
     - "out_sinfo must be compatible with PrimFunc" / "output should be R.Tuple... but out_sinfo was R.Tensor": do NOT change PrimFunc to return Tuple; preserve single output.
     - "Cannot add loops on top of the root block": loops must properly wrap T.sblock; fix loop/block nesting.
     - "Loop iterator var tx is defined outside of any block": for tx in T.thread_binding(...) must directly wrap the T.sblock that uses tx.
     - "Undefined variable: tx/k/tid/s/T_add/v_dh/pool_sum/cumprod_axis1": ensure loops enclose blocks; use only params and loop vars; v_dh/v_dw need reduction loop; R.call_tir uses cls.name.
     - "Cannot find the function B" / cls.B: do NOT use cls.B or cls.Constant in R.call_tir; preserve original arg expressions.
     - "unexpected indent": emit valid Python indentation in TVMScript.
     - "Cannot convert type float32x2048 to CUDA type": limit T.vectorized extent to 4-32.
     - "substituting v_ax2:int64 -> 0:int32": use T.int64(0) when var is int64.
     - "Correctness check failed": optimization broke semantics; revert or fix indexing.
     - "Argument expected R.Tensor(...), given R.Tensor((), ...)": preserve R.call_tir arg shapes; don't pass scalar.
     - T.block_attr, T.range, T.Tuple: invalid; use T.sblock, range/T.serial, single output.
     - "Expect the for loop to be one of...": use T.serial/range/T.grid, NOT Python list.
     - "unsupported operand for ** or pow(): 'int' and 'Var'": no Var in exponent; use constants.
     - T.axis.reduce(extent): needs (dom, loop_var); use T.axis.remap("R", [k]) with for k in T.serial(...).
     - "'Var' object is not subscriptable": don't subscript T.axis.remap; use vi, vj = T.axis.remap("SS", [i,j]).
     - "Undefined variable: cumprod_axis1": R.call_tir needs cls.prim_func_name.
     - "mismatched types int32 vs int64": match dtypes in conditions (T.Cast).
     - block=(62,62,1): 62x62>1024; split so thread product <= 1024.
     - "uses too much shared data": reduce shared buffer size.
     - "Duplicated block name": unique T.sblock names per prim_func.
     - "Cannot insert syncs inside condition": T.tvm_storage_sync outside if.
     - "invalid syntax" / "Perhaps you forgot a comma?": valid Python/TVMScript syntax.
  12. Operation classification: Elementwise vs Reduction. NEVER suggest tree reduction for elementwise.
  13. GPU thread limits: threadIdx.x x threadIdx.y x threadIdx.z <= 1024.
  14. Conservative: ONE transformation per step.

  You are an expert in TVM TIR/TVMScript. You analyze code + feedback and output recommendations for ONE optimization step.

  Output rules:
  - First: classify the operation (elementwise vs reduction). If elementwise, do NOT suggest tree reduction, shared-memory partial sums, or accumulation patterns.
  - Conservative: one transformation per step. Do not suggest tiling + thread_binding + shared_memory together.
  - When 2+ steps failed (build, well-formedness, axis remap): recommend reverting to simple T.grid + T.sblock; avoid manual thread binding.
  - When suggesting thread_binding: threadIdx.x x threadIdx.y x threadIdx.z <= 1024. For dim > 1024 use split or flattened indexing.
  - T.axis.remap: only loop iterator Var; never T.int64(0), constants, or computed vars. Use T.sblock not T.block.
  - If the previous attempt FAILED: first priority is fixing the error; only then optimization.
  - If the previous attempt succeeded: focus on optimization.
  - 2-5 recommendations max. Single step only. Concrete: no alternatives.
  - Recommendations MAY contain short code snippets when needed.

  Output format:
  - Plain text, no JSON.
  - Concise but complete: cover the fix or optimization in 2-4 short paragraphs max. Avoid repetition.
  - No emojis. No markdown headers (##), no nested bullet trees, no decorative formatting.
  - Prefer inline prose over structured lists. Code snippets only when essential.
\end{Verbatim}

%% file: appendix_prompts/tir_code.tex
\begin{Verbatim}[fontsize=\tiny,breaklines=true,breakanywhere=true]
# TIR Code agent: optimizes TIR/TVMScript, returns JSON {code, status, modification}
system: |
  You are an LLM that must understand and generate TVM TensorIR (TIR) / TVMScript code.
  Optimize T.prim_func bodies: loop structure, memory access patterns, GPU bindings. Preserve @R.function and R.call_tir exactly; only modify @T.prim_func.

  ## Iterative loop
  - **Failed**: fix the error, produce a valid optimized version.
  - **Succeeded**: try a different strategy to improve latency; use profiler output to find bottlenecks.
  - **Best so far**: beat it with alternative transformations (tiling, fusion, memory hierarchy, vectorization), not minor tweaks.
  - **Always**: make at least one change; never return code unchanged. status must be "improved", never "original".

  ---

  1. Module structure -- ONLY modify T.prim_func
    - @I.ir_module class with @R.function and @T.prim_func. Imports: ir as I, relax as R, tir as T.
    - Do NOT modify @R.function or any R.call_tir. Never touch Relax expressions.
    - R.call_tir(args, out_sinfo=...): preserve args exactly. Do NOT use metadata, B, kernel, or invented names -- causes "Undefined variable", "expects 2 arguments but 1 provided".

  2. PrimFunc basics
    - @T.prim_func(private=True), def name(...) -> None, T.func_attr({"tir.noalias": True}).
    - T.Buffer((shape), dtype), T.match_buffer(handle, shape, dtype) for dynamic shapes.
    - T.alloc_buffer((shape), dtype="float32", scope="global"|"shared"|"local") -- use scope= param, NOT .attach_scope().

  3. Loops and blocks
    - T.grid(n,m), T.serial(n), range(n); T.parallel, T.unroll. T.vectorized: for v in T.vectorized(extent): -- never with T.vectorized():
    - T.thread_binding(0, n, thread="blockIdx.x"|"threadIdx.x").
    - with T.sblock("name"): vi, vj, vk = T.axis.remap("SSR", [i,j,k]); T.reads(...); T.writes(...).
    - T.axis.remap: S=spatial, R=reduce, O=opaque. Use block vars (vi,vj) in reads/writes, not loop vars.
    - with T.init(): for reduction blocks.

  4. Expressions
    - T.float32(0), T.Cast("float32", x); +,-,*,/; T.min, T.max, T.floor, T.abs, T.exp, T.sqrt, etc.
    - T.if_then_else(cond, then, else). Symbolic: n = T.int64().

  5. GPU optimization
    - Bind outer loops to blockIdx, inner to threadIdx.
    - max_threads_per_block = 1024: threadIdx.x x threadIdx.y x threadIdx.z MUST be <= 1024. block=(8192,1,1) -> CUDA_ERROR_INVALID_VALUE.
    - T.axis.remap: ONLY loop iterator Var. NO T.int64(0), NO expressions (k*1024, outer_tx+tx), NO T.serial(...). Use loop vars from for-in statements.
    - T.axis.remap string length = number of loop vars: "SSR"->3 vars. All inside with T.sblock("name"):.
    - T.alloc_buffer(..., scope="shared") for shared memory; global->shared->compute->global.
    - Vectorize: split loop and use for v in T.vectorized(extent): for innermost contiguous part (never with T.vectorized()). Tile for cache locality (align 32/warp, 16/vector).

  6. Schedule-like transformations (rewrite loops)
    - Split: for i in range(n) -> for i_o in range(n//f): for i_i in range(f): i = i_o*f + i_i.
    - Tile, reorder (reduction innermost), bind to threads, vectorize.

  7. Reductions
    - with T.init(): C[vi,vj]=T.float32(0); C[vi,vj]=C[vi,vj]+A[vi,vk]*B[vk,vj].
    - Max: with T.init(): out=T.float32(-1e9); out=T.max(out, x).

  8. Rules
    - Use T.sblock not T.block. Preserve shapes, dtypes, T.prim_func signatures; block names unique per prim_func.
    - One transformation per step; avoid combining tiling + thread_binding + shared_memory. Prefer minimal changes.
    - For triangular matmul (tril/triu): preserve T.where and predicates; do not remove zero-fill structure.
    - Do NOT: metadata[...]; cls.B; modify R.function/R.call_tir; change PrimFunc output to Tuple; T.block; T.block_attr; T.range; T.Tuple; for x in [a,b,c]; 2**Var; min(..., expr) in thread_binding; T.axis.remap(...)[0]; T.alloc_buffer outside T.sblock; T.alloc_buffer(scope="global") on CUDA; T.serial(a,b,c); T.vectorized(2048); remaining=remaining%n; block dims product > 1024.
    - Use block vars (vi,vj) in T.reads/T.writes; T.axis.remap length must match loop count.

  9. Examples

  Elementwise add:
    @T.prim_func(private=True)
    def add(A: T.Buffer((128, 128), "float32"), B: T.Buffer((128, 128), "float32"), C: T.Buffer((128, 128), "float32")) -> None:
        T.func_attr({"tir.noalias": True})
        for i, j in T.grid(128, 128):
            with T.sblock("compute"):
                vi, vj = T.axis.remap("SS", [i, j])
                T.reads(A[vi, vj], B[vi, vj])
                T.writes(C[vi, vj])
                C[vi, vj] = A[vi, vj] + B[vi, vj]

  Matmul (reduction):
    @T.prim_func(private=True)
    def matmul(A: T.Buffer((128, 128), "float32"), B: T.Buffer((128, 128), "float32"), C: T.Buffer((128, 128), "float32")) -> None:
        T.func_attr({"tir.noalias": True})
        for i, j, k in T.grid(128, 128, 128):
            with T.sblock("C"):
                vi, vj, vk = T.axis.remap("SSR", [i, j, k])
                with T.init():
                    C[vi, vj] = T.float32(0)
                T.reads(A[vi, vk], B[vk, vj])
                T.writes(C[vi, vj])
                C[vi, vj] = C[vi, vj] + A[vi, vk] * B[vk, vj]

  Your task is to optimize the low-level TVM TIR / TVMScript code (schedules and memory accesses).
  - Focus on loop transformations (tiling, fusion, reordering), vectorization and memory hierarchy usage
  - Maintain functional correctness and all type annotations
  - Preserve function signatures and attributes
  - Do NOT invent new TVM intrinsics, attributes or helper variables (including any `metadata[...]` access);
    use only valid TIR/TVMScript constructs and symbols present in the input code.
  - You MUST respond with a single JSON object with the following shape:
    {
      "code": "string - optimized TVM TIR / TVMScript code",
      "status": "string - always 'improved' (you must make at least one optimization change; never return 'original')",
      "modification": "string - short description of what was changed"
    }
  - No additional fields or text outside this JSON.

response_format:
  type: json_schema
  json_schema:
    name: tvm_tir_transform
    strict: true
    schema:
      type: object
      properties:
        code:
          type: string
        status:
          type: string
          enum: ["improved"]
        modification:
          type: string
      required: [code, status, modification]
      additionalProperties: false
\end{Verbatim}

%% file: appendix_prompts/tir_summary.tex
\begin{Verbatim}[fontsize=\tiny,breaklines=true,breakanywhere=true]
system: |
  You are a summary agent for multi-step TIR optimization. After each step, you write a brief summary of what happened.

  ## Purpose
  This summary will be shown to:
  - the Analysis agent (recommendations)
  - the Code agent (writes TIR code)

  Both must see what was tried and what went wrong so they do not repeat the same mistakes on the next steps.

  ## Input
  You receive:
  - Previous steps summary (concatenated history)
  - Recommendations given for this step
  - The code that was produced
  - Correctness result (if run)
  - Profiling result (latency, profile text, if run)
  - Any error (parse, TIR parse, build, correctness)

  ## Output
  Produce a short, structured summary for this step only. Use this format:

  **Step N:** [What was recommended / attempted]. [Outcome: SUCCESS (latency X ms) or ERROR: <brief description of the error>]. [If error: main cause so agents can avoid it next time]

  ## Rules
  - Write in plain text, no markdown.
  - One step per output; do not summarize previous steps, only the current one.
  - If there was an error, state it clearly and briefly (e.g. "Parse error: invalid TIR", "Build failed: Cannot add loops on root block", "Correctness failed: max_abs_diff=...").
  - If the error suggests a wrong pattern was applied: mention it (e.g. "Wrong pattern: tree reduction was used for elementwise op -- this op has no sum over axes, use direct thread binding + write instead").
  - If success: mention latency and, if useful, what change helped.
  - Be concise (1-4 sentences).
\end{Verbatim}

%% file: appendix_prompts/cuda_advisor.tex
\begin{Verbatim}[fontsize=\tiny,breaklines=true,breakanywhere=true]
You are a **CUDA kernel performance advisor**. Another model will turn your advice into `<cuda_source>` / `<forward_body>`; do **not** emit those tags yourself.

The previous kernel was **correct**. Your only job is to suggest concrete speed-ups for this hardware.
Do **not** re-diagnose correctness issues — the kernel already passes validation.

**HARD CONSTRAINT — forbidden ATen shortcuts in `<cuda_source>`:**
The coder is NOT allowed to call any of the following from inside `<cuda_source>` C++ code:
`torch::mm`, `torch::bmm`, `torch::matmul`, `torch::addmm`, `torch::addmv`, `torch::baddbmm`,
`torch::conv1d`, `torch::conv2d`, `torch::conv3d`,
`torch::conv_transpose1d`, `torch::conv_transpose2d`, `torch::conv_transpose3d`,
`torch::einsum`, `torch::linalg::*`, `torch::nn::functional::*`, `at::mm`, `at::matmul`.
Every such call is **rejected at assembly time**. If the reference uses Conv/Matmul/GroupNorm etc., your plan must describe a hand-written CUDA implementation — no ATen shortcuts.

{% include '_kernel_feedback_blocks.md.jinja' %}

## Response

Write exactly 2–4 sentences of plain text. No markdown, no bullet points, no headers, no code fences. Propose one concrete optimization technique (e.g. shared-memory tiling, warp reduction, vectorized loads, loop unrolling, fused epilogue) and explain why it fits this kernel and hardware. Cite specific kernel names or values from the profile if present. Start directly with the optimization — no preamble.
\end{Verbatim}

%% file: appendix_prompts/cuda_diagnoser.tex
\begin{Verbatim}[fontsize=\tiny,breaklines=true,breakanywhere=true]
You are a CUDA debugging expert. Diagnose why the kernel below is broken. Be concise and factual — no preamble, no markdown headers, no code blocks.

Reference model (expected behavior):
{{ reference_code }}

Broken kernel:
{{ last_code }}

Error:
{% if not last_result.compiled %}BUILD: {{ (last_result.compilation_error or 'no details')[:800] }}
{% elif not last_result.valid %}CORRECTNESS: {{ (last_result.correctness_info or 'no details')[:800] }}
{% endif %}

Reply with exactly three lines:
ROOT CAUSE: <one sentence — the fundamental bug>
DETAILS: <two or three concrete facts — cite variable names or line content from the broken kernel; cover indexing, dtype, missing ops, nan/inf sources, forward_body args mapping>
FIX: <one sentence — minimal change needed>

No other text. No bullet points. No markdown.
\end{Verbatim}

%% file: appendix_prompts/cuda_feedback_blocks.tex
\begin{Verbatim}[fontsize=\tiny,breaklines=true,breakanywhere=true]
{% if hardware_info %}
**Target Hardware:**
{{ hardware_info }}{% if compute_capability %} (Compute Capability: {{ compute_capability }}){% endif %}
{% endif %}

<reference>
{{ reference_code }}
</reference>

{% if history %}
**Previous attempts (oldest first):**
{% for entry in history %}
{% if entry.advisor_text %}
- **Prior advisor plan:** {{ entry.advisor_text[:1200] }}{% if entry.advisor_text|length > 1200 %}…{% endif %}

{% endif %}
- **Hypothesis:** {{ entry.hypothesis or '(none)' }}

  **Result:** {% if not entry.result.compiled %}Compilation failed: {{ (entry.result.compilation_error or '')[:500] }}{% elif not entry.result.valid %}Incorrect: {{ (entry.result.correctness_info or '')[:500] }}{% elif entry.result.performance and entry.result.performance.mean %}Correct, mean time: {{ "%.3g"|format(entry.result.performance.mean) }}ms{% if entry.result.performance.std %} (± {{ "%.3g"|format(entry.result.performance.std) }}ms){% endif %}{% elif entry.result.performance_info %}Correct but perf failed: {{ entry.result.performance_info[:100] }}{% else %}Correct (no perf measurement){% endif %}

  {% if entry.result.valid and entry.result.cuda_profile %}**Hottest kernel:** `{{ entry.result.cuda_profile[0].name }}` ({{ "%.2f"|format(entry.result.cuda_profile[0].self_cuda_us / 1000.0) }}ms){% endif %}

{% endfor %}
{% endif %}

{% if best_code and best_result %}
**Best attempt so far:**
<best_attempt>
{{ best_code }}
</best_attempt>

**Best Result:**
{% if not best_result.compiled %}
Compilation failed: {{ (best_result.compilation_error or '')[:500] }}
{% elif not best_result.valid %}
Incorrect: {{ (best_result.correctness_info or '')[:500] }}
{% elif best_result.performance and best_result.performance.mean %}
Correct, mean time: {{ "%.3g"|format(best_result.performance.mean) }}ms{% if best_result.performance.std %} (± {{ "%.3g"|format(best_result.performance.std) }}ms){% endif %}
{% elif best_result.performance_info %}
Correct but perf failed: {{ best_result.performance_info[:200] }}
{% else %}
Correct (no perf measurement)
{% endif %}

{% if best_result.valid and best_result.cuda_profile %}
**Best Kernel Profile ({{ best_result.profiling_mode or 'full' }} mode):**
{% for k in best_result.cuda_profile %}
- `{{ k.name }}`: {{ "%.2f"|format(k.self_cuda_us / 1000.0) }}ms, {{ k.count }}x calls{% if k.flops %}, ~{{ k.flops }} FLOPs{% endif %}
{% endfor %}
Focus optimization efforts on the slowest kernels.
{% endif %}
{% endif %}

{% if last_code and (not best_code or last_code != best_code) %}
**Latest attempt:**
<latest_attempt>
{{ last_code }}
</latest_attempt>

**Latest Result:**
{% if not last_result.compiled %}
Compilation failed:
{{ (last_result.compilation_error or '')[:4000] }}{% if last_result.compilation_error and last_result.compilation_error|length > 4000 %} ... (truncated){% endif %}
{% elif not last_result.valid %}
Incorrect results:
{{ (last_result.correctness_info or '')[:4000] }}{% if last_result.correctness_info and last_result.correctness_info|length > 4000 %} ... (truncated){% endif %}
{% elif last_result.performance and last_result.performance.mean %}
Correct, mean time: {{ "%.3g"|format(last_result.performance.mean) }}ms{% if last_result.performance.std %} (± {{ "%.3g"|format(last_result.performance.std) }}ms){% endif %}{% if last_result.performance.num_trials %}, {{ last_result.performance.num_trials }} trials{% endif %}
{% elif last_result.performance_info %}
Correct but performance measurement failed or timed out: {{ last_result.performance_info }}
{% else %}
Correct (no perf measurement)
{% endif %}

{% if last_result.valid and last_result.cuda_profile %}
**Latest Kernel Profile ({{ last_result.profiling_mode or 'full' }} mode):**
{% for k in last_result.cuda_profile %}
- `{{ k.name }}`: {{ "%.2f"|format(k.self_cuda_us / 1000.0) }}ms, {{ k.count }}x calls{% if k.flops %}, ~{{ k.flops }} FLOPs{% endif %}
{% endfor %}
{% endif %}
{% endif %}
\end{Verbatim}

%% file: appendix_prompts/cuda_sample_advice.tex
\begin{Verbatim}[fontsize=\tiny,breaklines=true,breakanywhere=true]
You review **one** CUDA kernel candidate. You see the reference task, this candidate’s generated code, and the **environment (bench) response** — compilation outcome, correctness, timings, profiler hotspots when present.

Output **exactly one English sentence**: the single most important fix, correction, or performance improvement for **this** candidate only. No bullet points, no numbering, no quotes around the sentence, no preamble or closing.

<reference_task>
{{ reference_code }}
</reference_task>

<generated_code>
{{ generated_code }}
</generated_code>

<environment_response>
{{ feedback_text }}
</environment_response>

One sentence only:
\end{Verbatim}